\pdfoutput=1
\documentclass[11pt]{article}

\usepackage[]{EMNLP2023}

\usepackage{times}
\usepackage{latexsym}

\usepackage{booktabs}
\usepackage[inline]{enumitem}
\usepackage{hyperref}       % hyperlinks
\usepackage{url}            % simple URL typesetting
\usepackage{amsfonts}       % blackboard math symbols
\usepackage{nicefrac}       % compact symbols for 1/2, etc.
\usepackage{microtype}      % microtypography

\usepackage{amsfonts}
\usepackage{amsmath}
\usepackage{multirow}
\usepackage{graphicx}
\usepackage{enumitem}
\usepackage{subcaption}
\usepackage{caption}
\usepackage{rotating}
\usepackage[normalem]{ulem}
\usepackage{amssymb}
\usepackage{colortbl}
\usepackage{linguex}

\usepackage{siunitx} 
\usepackage{multirow, makecell}
\usepackage{pifont}% http://ctan.org/pkg/pifont
\usepackage{siunitx} 
\usepackage{multirow, makecell}
\usepackage{tabularx}% added for table design
\usepackage[all]{nowidow}
\usepackage{colortbl}
\usepackage{nicematrix,tikz}
\definecolor{darkgreen}{rgb}{0.0, 0.42, 0.24}
\definecolor{green}{RGB}{112, 173,71}
\definecolor{blue}{RGB}{68, 114,196}
\definecolor{orange}{RGB}{237, 125,49}
\definecolor{red}{RGB}{202, 54,49}
\definecolor{yellow}{RGB}{222,194, 142}

\definecolor{hotpink}{HTML}{EF7C8E}
\definecolor{tiffanyblue}{HTML}{A0E7E5}
\definecolor{mint}{HTML}{B4F8C8}
\definecolor{paleyellow}{HTML}{FBE7C6}
\definecolor{rosewater}{HTML}{D8A7B1}
\definecolor{cream}{HTML}{FAE8E0}

\newcommand{\ts}{\textsc{TS}\xspace}

\newcommand{\asset}{\textsc{ASSET}\xspace}
\newcommand{\medeasi}{\textsc{Med-EASi}\xspace}
\newcommand{\newsela}{\textsc{Newsela}\xspace}

\usepackage[T1]{fontenc}
\usepackage[utf8]{inputenc}

\usepackage{microtype}

\usepackage{inconsolata}

\usepackage{tabularx}
\usepackage{caption}
\usepackage{subcaption}
\usepackage{tikz}
\usetikzlibrary{shapes.misc, positioning, decorations.pathreplacing, calc}

\usepackage{xargs}                      % Use more than one optional parameter in a new commands
\usepackage[colorinlistoftodos,prependcaption,textsize=small]{todonotes}
\newcommandx{\plan}[2][1=]{\todo[inline,linecolor=yellow,backgroundcolor=yellow!25,bordercolor=yellow,#1]{TODO: #2}}
\newcommandx{\unsure}[2][1=]{\todo[linecolor=red,backgroundcolor=red!25,bordercolor=red,#1]{#2}}
\newcommandx{\change}[2][1=]{\todo[linecolor=blue,backgroundcolor=blue!25,bordercolor=blue,#1]{#2}}
\newcommandx{\info}[2][1=]
{\todo[linecolor=OliveGreen,backgroundcolor=OliveGreen!25,bordercolor=OliveGreen,#1]{#2}}
\newcommandx{\improvement}[2][1=]
\title{BLESS: Benchmarking Large Language Models on Sentence Simplification}

\author{Tannon Kew$^{1,\dagger}$, Alison Chi$^{2,\dagger}$, Laura Vásquez-Rodríguez$^{3,4,\dagger,\ast}$, \\
\textbf{Sweta Agrawal}$^5$, \textbf{Dennis Aumiller}$^6$, \textbf{Fernando Alva-Manchego}$^7$, \textbf{Matthew Shardlow}$^8$ \\
  $^1$University of Zurich, Switzerland 
  $^2$National Tsing Hua University, Taiwan \\
  $^3$Idiap Research Institute, Switzerland 
  $^4$University of Manchester, UK \\
  $^5$University of Maryland, US
  $^6$Cohere, US 
  $^7$Cardiff University, UK \\
  $^8$Manchester Metropolitan University, UK \\
  \texttt{kew@cl.uzh.ch, achi@gapp.nthu.edu.tw, laura.vasquez@idiap.ch} \\
  \texttt{sweagraw@umd.edu, dennisaumiller@cohere.com} \\
  \texttt{alvamanchegof@cardiff.ac.uk, m.shardlow@mmu.ac.uk} \\
  \\}

\begin{document}
\maketitle
\begin{abstract}

We present \textbf{BLESS}, a comprehensive performance benchmark of the most recent state-of-the-art large language models (LLMs) on the task of text simplification (\ts). 
We examine how well off-the-shelf LLMs can solve this challenging task, assessing a total of 44 models, differing in size, architecture, pre-training methods, and accessibility, on three test sets from different domains (Wikipedia, news, and medical) under a few-shot setting. 
Our analysis considers a suite of automatic metrics as well as a large-scale quantitative investigation into the types of common edit operations performed by the different models. 
Furthermore, we perform a manual qualitative analysis on a subset of model outputs to better gauge the quality of the generated simplifications.
Our evaluation indicates that the best LLMs, despite not being trained on \ts, perform comparably with state-of-the-art \ts baselines.
Additionally, we find that certain LLMs demonstrate a greater range and diversity of edit operations.
Our performance benchmark will be available as a resource for the development of future \ts methods and evaluation metrics.\footnote{We make our code and the generated system outputs available at \url{https://github.com/ZurichNLP/BLESS}.}
\def\thefootnote{†}\footnotetext{These authors contributed equally.}
\def\thefootnote{*}\footnotetext{Work done as a PhD student at the University of Manchester, United Kingdom.} 

% Our findings reveal that although the best LLMs fall short of state-of-the-art TS baselines according to automatic metrics, they perform more desirable edit operations. This highlights the need for automatic metrics that better reflect the goals of TS.

% Previous work has focused on the evaluation of \ts on a handful of selected models (such as ChatGPT) which, despite not being explicitly trained for the task, has shown to surpass the simplification abilities of supervised TS models. But there is still no large-scale and detailed analysis of the simplification capabilities of LLMs due to the resources and costs required to perform such an evaluation. 

\end{abstract}

\section{Introduction}

Large pre-trained language models (LLMs) have demonstrated strong performance on a wide range of NLP tasks without the need for task-specific fine-tuning, leading to a prevailing conventional wisdom that LLMs can solve \textit{any} task. This has motivated the development of benchmarks to better understand the abilities of LLMs in specific domains such as healthcare \citep{Malik_2023}, finance \citep{Dowling_2023}, education \citep{baidoo2023education}, engineering \citep{sobania2023analysis}, and ethics \citep{zhuo2023red}, as well as for specific NLP tasks \cite{li-etal-2022-eliciting, wang2023documentlevel, liu2023geval}.

However, it remains unclear how well current LLMs can perform on the challenging task of text simplification (\ts).
% , which aims reduce the linguistic complexity of a text in order to make it easier to understand and more accessible.
%For our English language benchmark, we mostly focus on sentence simplification
In this paper, we focus on sentence simplification in English, which typically involves rephrasing part or all of a sentence into language which is more accessible and easier to understand.
While recent work has focused on evaluating \ts abilities of select models, such as \texttt{GPT-3.5-Turbo} \citep{feng_sentence_2023} and \texttt{mT5} \citep{ryan-etal-2023-revisiting}, there is currently no large-scale and detailed analysis of the simplification capabilities of different LLMs. % due to the resources and costs required to perform such an evaluation.

In this study, we expand both the breadth and depth of the knowledge base on \ts with LLMs, evaluating a wider variety of models on three different \ts datasets: \asset \citep{alva-manchego-etal-2020-asset}, \newsela \citep{jiang-etal-2020-neural} and \medeasi \citep{basu2023medeasi}. 
We select these datasets to cover a variety of domains (Wikipedia, news, and medical) and a diverse set of \ts operations (e.g. paraphrasing, splitting, and elaboration).

Specifically, we use in-context learning (ICL) and assess LLMs in a few-shot setting, experimenting with three different prompts. 
We select 44 widely used generative models (both open and closed-weight) and evaluate their abilities from three distinct angles.
%To measure performance, we primarily rely on automatic metrics but also aim to quantify the nature of TS edit operations performed by the different models, and perform a targeted manual qualitative analysis to better gauge the quality of the generated simplifications.
% We evaluate the LLMs' abilities to simplify text . 
First, we rely on automatic evaluation metrics commonly used in the \ts literature. 
Second, we quantify and compare the edit operations performed by the LLMs during simplification. 
Finally, we perform a targeted qualitative analysis to validate our findings and to better understand the quality of the generated simplifications.
% \improvement{Add a paragraph summarising the main findings}
Our findings reveal that closed-weight models provide significant gains over open-weight alternatives under a few-shot setting, establishing them as a strong baseline for future work on \ts.
We summarize our contributions as follows:
    \begin{enumerate}[noitemsep,nolistsep]
        \item BLESS (\textbf{B}enchmarking \textbf{L}arge language mod\textbf{E}ls on \textbf{S}entence \textbf{S}implification), a performance evaluation benchmark of 44 LLMs in a few-shot setting (Section \ref{sec:methods}).
        \item An evaluation that includes both widely used automatic metrics and an analysis of the \ts edit operations performed by the models (Section \ref{sec:automatic_evaluation}).
        \item A qualitative analysis of the results, with manual annotation of simplification operations and an examination of the relationships between selected evaluation metrics (Section \ref{sec:qualitative_analysis}).
    \end{enumerate}

\section{Related Work}
%Previous work has created simplification benchmarks, investigated LLM simplification, and created methods for simplification evaluation.

\paragraph{Text Simplification Benchmarks} Most simplification work treats the task as a monolingual machine translation problem, training models on datasets containing complex-simple sentence pairs \citep{zhu-etal-2010-monolingual}. \citet{alva-manchego-etal-2020-data} performed a standardized evaluation of general data-driven simplification systems, using Wikipedia-based datasets and \newsela. 
At the document level, \citet{alva-manchego-etal-2019-cross} conducted a systematic analysis of simplification operations to demonstrate the limitations and disruptions that occur when multiple sentences are involved. 
Benchmarks have also been established for more specific kinds of simplification: for example, both non-neural \citep{paetzold-specia-2016-benchmarking} and neural \citep{Sanja_2022, saggion-etal-2022-findings} approaches to lexical simplification, which aims to replace complex words with simpler alternatives. 
% While most work has been focused on English, there are also some benchmark datasets for languages such as German \citep{sauberli-etal-2020-benchmarking} and Spanish \citep{gonzalez-dios-etal-2022-irekialfes}. 
% \improvement{How are these benchrmarks 'related' to ours? State what we are doing similar/different, or building from them.} 

\paragraph{LLM-based Simplification} LLMs such as \texttt{GPT-3.5-Turbo}, the model behind early versions of ChatGPT\footnote{\url{https://chat.openai.com/}}, are often used out-of-the-box without any further training for a given domain or task. 
Some previous works have investigated simplification capabilities of select LLMs in order to benchmark performance against dedicated approaches \citep{aumiller-gertz-2022-unihd,vasquez-rodriguez-etal-2022-uom, ryan-etal-2023-revisiting,sun-etal-2023-teaching, chi2023learning}. 
Meanwhile, \citet{feng_sentence_2023} explored the \ts abilities of the two strong-performing OpenAI models, \texttt{GPT-3.5-Turbo} and \texttt{Davinci-003}.
% Some of these works have proposed sentence simplification benchmarks for English, Spanish and Portuguese \cite{feng_sentence_2023} and in a multilingual setting \cite{ryan-etal-2023-revisiting} which suggests its great potential for \ts. 
However, despite these efforts, we only have results from a very limited number of LLMs and evaluation metrics. 
Thus, it remains unclear how a wider spectrum of models, differing in architecture and training strategy, perform on different domains and in response to different prompts.  
We aim to fill this gap and study the simplification abilities of 44 LLMs in order to highlight potential weaknesses and determine areas for further development.
To the best of our knowledge, we are the first to focus on establishing the performance of recent LLMs on the task of \ts.

\section{BLESS: Benchmarking Large Language Models on Sentence Simplification}
\label{sec:methods}

\begin{table}[t]
\centering
 \def\arraystretch{1.2}
\scalebox{0.72}{
\begin{tabular}{@{}lllrrrr@{}}
\toprule
% \rowcolor{gray!24}
\textbf{Dataset} & \textbf{Domain} & \textbf{Size} & \multicolumn{2}{c}{\textbf{\# Words}}   & \multicolumn{1}{c}{\textbf{\# R}} & \multicolumn{1}{c}{\textbf{TER}}   \\
% \rowcolor{gray!16}
       &   & &  \multicolumn{1}{c}{C} &  \multicolumn{1}{c}{S} &  &  \\
\midrule
 \textsc{\asset} & Wikipedia & 359 & 22.57 & 18.87 & 10 & 16.79\\
 \textsc{\medeasi} & Medical & 300 & 26.48 & 27.42 & 1 & 25.03\\ 
 \textsc{\newsela} &News & 256 & 26.44 & 24.82  & 4 & 23.17\\
 \bottomrule
  \end{tabular}
}
\caption{Dataset Statistics. C: Complex; S: Simple; R: References. TER refers to Translation Error Rate, a measurement of the average edit distance between the source and reference texts (see \url{https://www.cs.umd.edu/~snover/tercom}).} \label{tab:data_stats}

\end{table}

\subsection{Datasets}
Our assessment establishes the performance of current LLMs on \ts according to three datasets, covering different domains and styles. Table \ref{tab:data_stats} summarizes these datasets.

\textbf{\asset} \cite{alva-manchego-etal-2020-asset} comprises 2,359 sentences from English Wikipedia paired with 10 simplified references. We use the official test split (359 sentences) for evaluation. 
These references were created by crowdworkers who were instructed to use edit operations such as replacement, splitting, and deletion. 
% (see Figure \ref{fig:prompt2}).

\textbf{\medeasi} \citep{basu2023medeasi} is a simplification dataset for short medical texts containing 1,979 complex (expert) - simple (layman) pairs. Each text contains one or more sentences.
In this dataset, simplified texts are composed using four types of operations: elaboration, replacement, deletion, and insertion. 
We use the released test split (300 instances) for our evaluation.
Unlike the other two datasets, simplifications in \medeasi are slightly longer than the complex source texts, due to explanation and decomposition of complex medical terms.

\textbf{\newsela} \cite{xu-etal-2015-problems} contains 1,130 long-form news articles that have been professionally rewritten according to four different graded readability levels. 
% For our experiments, we did not perform any preprocessing steps from the original sources of \asset and \medeasi as described in Section \ref{sec:benchmark}. 
For our benchmarking experiments, we opt for the Newsela-Manual test set \cite{jiang-etal-2020-neural}.
We extract all aligned and partially aligned sentence pairs between a complex source sentence (level 0) and the four simplified article versions (levels 1-4), keeping only those sentences for which we have a reference for all four simplification levels.\footnote{These articles are simplified as a whole to match the desired school grade; therefore, there is no guarantee that there will be an exact match for all the sentences in the text across all grade levels.}
This results in 256 test examples. 
Using this small subset of \newsela data ensures that sentence-level alignments are of high quality and capture important edit operations such as splitting.

% \improvement{We could add a paragraph summarizing/highlighting what the concrete differences between these datasets allow us to compare. For example, the differences in vocabulary, the length of the input texts, the single vs multi operations, etc. Basically, why of all available datasets, we decided our benchmark to have these specific ones?}

% \subsection{Models}

%\plan{Introduction to the simplification models, how did we choose them?}

% \plan{Characterisation of the benchmark. Add statistics/example from each of the dataset}
\subsection{LLM Types}
We investigate a total of 44 LLMs with different sizes, architectures, and training objectives.
The models we consider range from 60 million to 176 billion parameters and are all based on the transformer architecture \citep{vaswani2017}, consisting of either an encoder-decoder or a standalone decoder. 
Furthermore, all have undergone a self-supervised pre-training stage.
%we broadly distinguish between models where weights are accessible ("open-weight"), and commercial models only accessible through API services ("closed-weight").
Nine of these models leverage instruction-tuning, which fine-tunes a pre-trained base model on labeled instruction-response pairs from a diverse set of tasks.
Finally, just three of these models have received additional training through reinforcement learning with human feedback (RLHF) to better align the model's responses with human preferences \citep{stiennon_learning_2020, ouyang2022training}.
Evaluating a wide variety of currently available models should serve as a broad baseline and give sufficient information on which models perform best in which domains as well as where key challenges remain.
%\unsure{reinforce that comparing pre-training objectives is not our goal, but instead to benchmark a wide array of currently available LMs?}

We broadly distinguish between open- and closed-weight models.
The former pertains to models for which the trained weights are accessible and thus allow for self-hosting. Typically, these models are considered to be ``open-source.'' However, we note that this obfuscates specific licensing agreements attached to some models and whether or not the training data and code are also made available.
In comparison, closed-weight models refer to those whose weights are kept private and can be queried only through APIs. 
Our open-weight models include variants of the \texttt{T5} family \citep{raffel2020exploring}, \texttt{GPT}-style models \citep{radford2019language, gpt-j}, \texttt{OPT} \cite{zhang2022opt} and \texttt{LLaMA} models \citep{touvron2023llama}, and the \texttt{BLOOM} family \citep{scao2022bloom}. 
For closed-weight models, we focus on those developed by OpenAI. Details on each model family are provided in Appendix \ref{sec:appendix_model_details}.
% Typically, closed-weight models are less
% for which no weights are available.

% \plan{Maybe a table that summarise their most important features? (Dennis: I think it would be a great idea, potentially for an Appendix, to list the different training objectives, or compare the pre-training/parameter size)}

%\plan{Description of each simplification model (probably this will end in the appendix) analysis}

%\plan{Consider grouping the individual subsections together. E.g., T5/T0/Flan/UL2 all have the same "architecture stack", and mainly differ in some fringe pre-training objective. Still, explanations could be given in a single section, which abbreviates explanation. Same goes for OPT/LLaMA/BLOOM}

% \usepackage{caption}
% \usepackage{subcaption}
% \usepackage{tikz}
% \usetikzlibrary{shapes.misc, positioning, decorations.pathreplacing, calc}

%%%%%%%%%%%%%%%%%%%%%%%%%%%%%%%%
%%%%% Begin Prompt figures %%%%%
%%%%%%%%%%%%%%%%%%%%%%%%%%%%%%%%

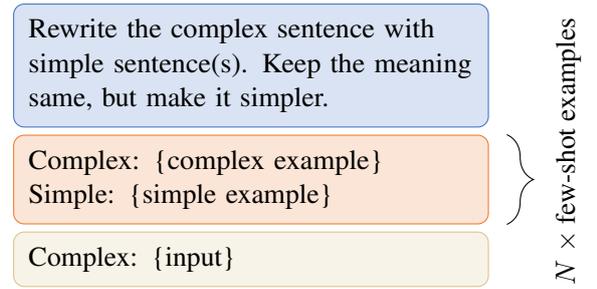
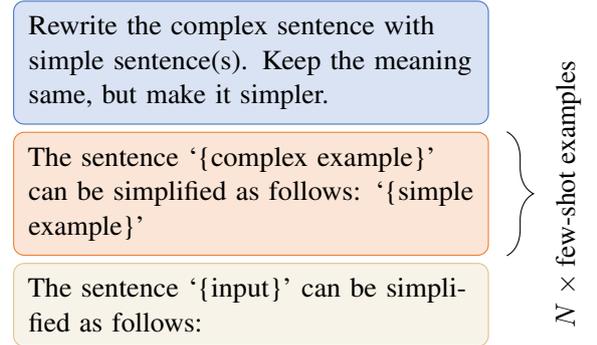
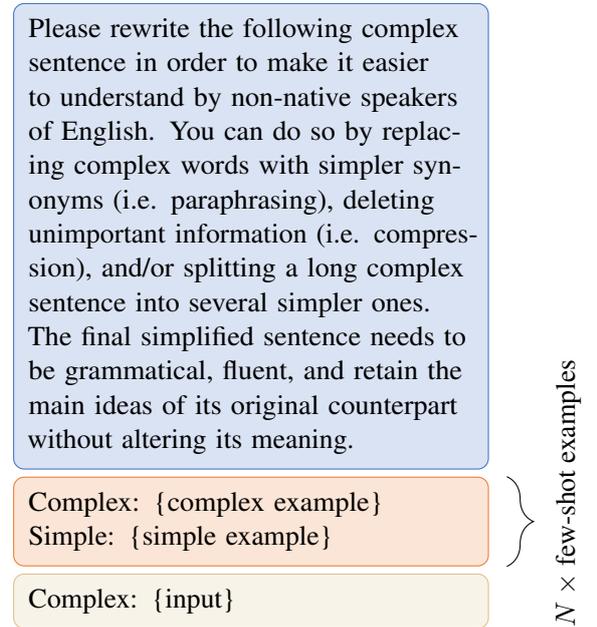
\begin{figure}[htb!]

\begin{subfigure}[b]{\linewidth}
\centering
\begin{tikzpicture}[scale=0.95, every node/.style={transform shape}]

% Blue rectangle for "Prompt"
\node[rectangle, rounded corners, draw=blue, fill=blue!20, text width=0.8\linewidth, align=left, inner sep=1.2ex] (prompt) {Rewrite the complex sentence with simple sentence(s). Keep the meaning same, but make it simpler.};

% Orange rectangle for "Examples"
\node[rectangle, rounded corners, draw=orange, fill=orange!20, below=0.1cm of prompt, text width=0.8\linewidth, align=left, inner sep=1.2ex] (examples) {
Complex: \{complex example\} 
\linebreak 
Simple: \{simple example\}
};

% Yellow rectangle for "Input"
\node[rectangle, rounded corners, draw=yellow, fill=yellow!20, below=0.1cm of examples, text width=0.8\linewidth, align=left, inner sep=1.2ex] (input) {Complex: \{input\}};

% Add curly brace
\draw [decorate,decoration={brace,amplitude=10pt,mirror,raise=4pt},yshift=0pt]
($(examples.south east)+(0.1cm,0)$) -- ($(examples.north east)+(0.1cm,0)$) node [black,midway, yshift=0.4cm, xshift=1cm, rotate=90] {$N$ $\times$ few-shot examples};

\end{tikzpicture}
    \caption{Prompt 0 uses a basic instruction adapted from \cite{feng_sentence_2023} followed by a list of $N$ few-shot examples before the input sentence to be simplified.}
    \label{fig:prompt0}
% \end{figure}
\end{subfigure}

% \begin{figure}
\begin{subfigure}[b]{\linewidth}

\vspace{0.2cm}

\centering
\begin{tikzpicture}[scale=0.95, every node/.style={transform shape}]

% Blue rectangle for "Prompt"
\node[rectangle, rounded corners, draw=blue, fill=blue!20, text width=0.8\linewidth, align=left, inner sep=1.2ex] (prompt) {Rewrite the complex sentence with simple sentence(s). Keep the meaning same, but make it simpler.};

% Orange rectangle for "Examples"
\node[rectangle, rounded corners, draw=orange, fill=orange!20, below=0.1cm of prompt, text width=0.8\linewidth, align=left, inner sep=1.2ex] (examples) {
The sentence `\{complex example\}' can be simplified as follows: `\{simple example\}'
};

% Yellow rectangle for "Input"
\node[rectangle, rounded corners, draw=yellow, fill=yellow!20, below=0.1cm of examples, text width=0.8\linewidth, align=left, inner sep=1.2ex] (input) {The sentence `\{input\}' can be simplified as follows:};

% Add curly brace
\draw [decorate,decoration={brace,amplitude=10pt,mirror,raise=4pt},yshift=0pt]
($(examples.south east)+(0.1cm,0)$) -- ($(examples.north east)+(0.1cm,0)$) node [black,midway,xshift=1cm, rotate=90] {$N$ $\times$ few-shot examples};

\end{tikzpicture}
    \caption{Prompt 1 uses the same basic task instruction as prompt 0, but presents few-shot examples in an inline, continuous text format.}
    \label{fig:prompt1}
% \end{figure}
\end{subfigure}

\vspace{0.2cm}

% \begin{figure}
\begin{subfigure}[b]{\linewidth}
\centering
\begin{tikzpicture}[scale=0.95, every node/.style={transform shape}]

% Blue rectangle for "Prompt"
\node[rectangle, rounded corners, draw=blue, fill=blue!20, text width=0.8\linewidth, align=left, inner sep=1.2ex] (prompt) {Please rewrite the following complex sentence in order to make it easier to understand by non-native speakers of English. You can do so by replacing complex words with simpler synonyms (i.e. paraphrasing), deleting unimportant information (i.e. compression), and/or splitting a long complex sentence into several simpler ones. The final simplified sentence needs to be grammatical, fluent, and retain the main ideas of its original counterpart without altering its meaning.};

% Orange rectangle for "Examples"
\node[rectangle, rounded corners, draw=orange, fill=orange!20, below=0.1cm of prompt, text width=0.8\linewidth, align=left, inner sep=1.2ex] (examples) {
Complex: \{complex example\} 
\linebreak 
Simple: \{simple example\}
};

% Yellow rectangle for "Input"
\node[rectangle, rounded corners, draw=yellow, fill=yellow!20, below=0.1cm of examples, text width=0.8\linewidth, align=left, inner sep=1.2ex] (input) {Complex: \{input\}};

% Add curly brace
\draw [decorate,decoration={brace,amplitude=10pt,mirror,raise=4pt},yshift=0pt]
($(examples.south east)+(0.1cm,0)$) -- ($(examples.north east)+(0.1cm,0)$) node [black,midway, yshift=0.4cm, xshift=1cm, rotate=90] {$N$ $\times$ few-shot examples};

\end{tikzpicture}
    \caption{Prompt 2 repurposes the instructions from \cite{alva-manchego-etal-2020-asset} that were provided to crowdworkers in the creation of the ASSET dataset. Similarly to prompt 0, few-shot examples are presented in a structured format.}
    \label{fig:prompt2}
\end{subfigure}
\caption{Prompts used for LLM text simplification. The blue boxes contain the task instructions. Orange boxes show how the few-shot examples are presented to the model and yellow boxes contain the prefix for the model to continue.}
\label{fig:prompts}

\end{figure}

%%%%%%%%%%%%%%%%%%%%%%%%%%%%%%
%%%%% End Prompt figures %%%%%
%%%%%%%%%%%%%%%%%%%%%%%%%%%%%%
% \section{Experiments}

\subsection{Prompts}
To simplify sentences with LLMs without additional fine-tuning, we use in-context learning (ICL). ICL is a prompting technique that utilizes a small number of input-output examples to demonstrate a task \cite{brown2020language}.
Previous work on related tasks has demonstrated that LLMs are sensitive to which input prompts and few-shot examples are used \cite{zhang-etal-2022-prompt, lu_fantastically_2022, agrawal-etal-2023-context}. 
To account for this, we construct three stylistically distinct prompts that consist of a task instruction and $N$ few-shot examples (see Figure \ref{fig:prompts}).
For all generation settings, we set $N$=3 and randomly sample complex-source pairs from the corresponding validation sets.
We leave a detailed investigation of optimal in-context learning strategies for \ts to future work.

\subsection{Inference Settings}

For open-weight models, we run inference on local GPUs using the Transformers library~\citep{wolf-etal-2020-transformers}. We load the models with 8-bit quantization \cite{dettmers_llmint8_2022}, which allows us to run inference efficiently on as few as 5 A100 80GB GPUs.
For closed-weight models, we use the APIs provided by OpenAI.
% and Cohere.
% Naturally, local inference allows for more customisation of generation hyperparameters.
As generation hyperparameters, we use Nucleus Sampling \cite{holtzman_curious_2020} with a probability threshold of 0.9, a temperature of 1.0, and a maximum output length of 100 tokens.
To account for the stochastic generation settings, we perform each inference run with 3 different random seeds and aggregate the results for each metric. 
% For more details regarding the hardware used, see Appendix \ref{sec:app:hardware}.

\subsection{Baselines}

We use the MUSS \cite{martin-etal-2022-muss} model as our main baseline since it has been shown to achieve state-of-the-art performance. MUSS fine-tunes a BART-large \cite{lewis-etal-2020-bart} model with ACCESS control tokens \cite{martin-etal-2020-controllable} extracted from labeled \ts datasets and/or mined paraphrases to train both supervised (\texttt{MUSS-wiki-mined}) and unsupervised (\texttt{MUSS-mined}) \ts systems. We use the suggested hyperparameters from the original paper to set the control tokens for simplification generation. 

\subsection{Automatic Metrics} \label{sec:automatic_metrics}
To assess how well LLMs can perform \ts, we evaluate all the model outputs using a suite of automatic metrics.\footnote{See Appendix \ref{sec:sup_eval_metrics} for details on each evaluation metric.} 
We measure simplicity using SARI \citep{xu-etal-2016-optimizing}, meaning preservation using BERTScore \citep{Zhang_2020}, and readability using FKGL \citep{Kincaid-1975}.
These metrics are computed using the EASSE package \citep{alva-manchego-etal-2019-easse}.\footnote{\url{https://github.com/feralvam/easse}}
Additionally, we report LENS \cite{maddela-etal-2023-lens}, a recently proposed learned metric, which considers both the semantic similarity and the degree of simplification performed by the system with respect to the source sentence and references.\footnote{We compute LENS using its original implementation: \url{https://github.com/Yao-Dou/LENS}.}
Where possible, we also establish the performance of `gold' simplifications by evaluating available reference sentences using a `leave-one-out' strategy. That is, in cases where multiple references are available, we select one at random and evaluate it against the remaining references.

\section{Automatic Evaluation Results} \label{sec:automatic_evaluation}

In this section, we present the results of our automatic evaluation of simplification outputs and summarize our main findings.
First, we perform an exhaustive assessment using automatic metrics (Section \ref{sec:automatic_metrics}). 
For brevity, we report the results of the best-performing LLMs with SARI and BERTScore in Table \ref{table:results-summary} and provide the complete results for all 44 models and metrics in Appendix \ref{sec:appendix_more_results}.
Then, we compute edit distance statistics to quantify the simplification operations performed by each of the LLMs (Section \ref{sec:quantitative_analysis}).
We begin by assessing the impact of the different prompt formats.

% on the three \ts test sets under the different few-shot prompt strategies adopted. 
% Additionally, we include a summary of the best-performing models on the \asset dataset in Table \ref{table:results-summary}.
% First off, we assess the impact of the different prompt formats used the various model sizes on simplification.
% We present results for each dataset according to automatic metrics and quantify the edit operations performed.

\begin{figure*}[h]
\centering
\includegraphics[width=0.90\linewidth]{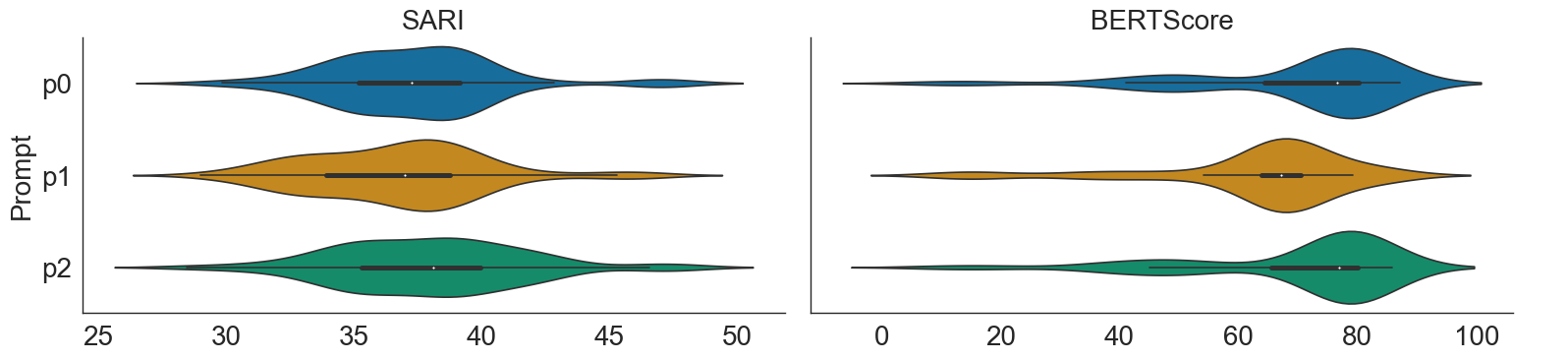}
\caption{Impact of prompt selection on SARI and BERTScore for all models on \asset. Prompts 0 and 2 achieve improved meaning preservation over prompt 1.}
\label{fig:all_prompts_violin}
\end{figure*}

\begin{table*}[htb!]
\centering
\label{table:asset-results-summary} 
\resizebox{0.9\textwidth}{!}{%
\begin{tabular}{@{}llcccccc@{}}
\toprule
&   & \multicolumn{2}{c}{\textbf{\asset}} & \multicolumn{2}{c}{\textbf{\medeasi}} & \multicolumn{2}{c}{\textbf{\newsela}} \\
& & \textbf{SARI $\uparrow$} & \textbf{BERT$\uparrow$} & \textbf{SARI$\uparrow$} & \textbf{BERT$\uparrow$} & \textbf{SARI$\uparrow$}                 & \textbf{BERT$\uparrow$} \\
\midrule %\midrule
\multirow{3}{*}{\textbf{Baselines}} & Gold References         & 45.27                         & 78.89                         & 100                           & 100                           & 60.11                         & 87.66                         \\
& \texttt{MUSS-mined}              & \cellcolor[HTML]{88CFAD}42.29 & \cellcolor[HTML]{65B3C1}79.86 & \cellcolor[HTML]{C7E9D8}35.15 & \cellcolor[HTML]{5DAFBE}42.55 & \cellcolor[HTML]{7DCBA5}38.40  & \cellcolor[HTML]{5FB0BF}72.14 \\
& \texttt{MUSS-wiki-mined}         & \cellcolor[HTML]{71C69C}44.90 & \cellcolor[HTML]{6AB5C3}77.71 & \cellcolor[HTML]{C8E9D9}35.12 & \cellcolor[HTML]{5BAEBD}43.07 & \cellcolor[HTML]{57BB8A}41.24 & \cellcolor[HTML]{5BAEBD}74.1  \\
\midrule % \midrule \midrule
% \textbf{LLMs}                    & \multicolumn{1}{l}{}          & \multicolumn{1}{l}{}          & \multicolumn{1}{l}{}          & \multicolumn{1}{l}{}          & \multicolumn{1}{l}{}          & \multicolumn{1}{l}{}          \\
% \midrule % \midrule 
\multirow{18}{*}{\textbf{LLMs}} & \texttt{Ada-001*}     & \cellcolor[HTML]{D4EEE1}33.97 & \cellcolor[HTML]{61B1BF}81.76 & \cellcolor[HTML]{A8DCC3}36.52 & \cellcolor[HTML]{7BBECA}33.95 & \cellcolor[HTML]{B2E0C9}34.42 & \cellcolor[HTML]{64B3C0}70.33 \\
 & \texttt{Babbage-001*} & \cellcolor[HTML]{ABDDC5}38.44 & \cellcolor[HTML]{5FB0BF}82.46 & \cellcolor[HTML]{A6DBC2}36.6  & \cellcolor[HTML]{6DB7C4}37.95 & \cellcolor[HTML]{97D5B7}36.41 & \cellcolor[HTML]{75BBC8}62.99 \\
 & \texttt{Curie-001*}   & \cellcolor[HTML]{9ED8BC}39.87 & \cellcolor[HTML]{5FB0BE}82.75 & \cellcolor[HTML]{82CDA8}38.22 & \cellcolor[HTML]{68B5C2}39.31 & \cellcolor[HTML]{89CFAD}37.53 & \cellcolor[HTML]{67B4C2}69.17 \\
 & \texttt{Davinci-002*} & \cellcolor[HTML]{83CDA9}42.84 & \cellcolor[HTML]{57ACBB}85.91 & \cellcolor[HTML]{ACDEC6}36.34 & \cellcolor[HTML]{59ADBC}43.67 & \cellcolor[HTML]{65C194}40.25 & \cellcolor[HTML]{5CAFBD}73.62 \\
 & \texttt{Davinci-003*} & \cellcolor[HTML]{61BF91}46.60 & \cellcolor[HTML]{65B3C1}79.66 & \cellcolor[HTML]{5FBE90}39.81 & \cellcolor[HTML]{63B2C0}40.83 & \cellcolor[HTML]{86CEAB}37.76 & \cellcolor[HTML]{79BDC9}61.56 \\
 & \texttt{GPT-3.5-Turbo*}   & \cellcolor[HTML]{57BB8A}47.69 & \cellcolor[HTML]{66B4C1}79.39 & \cellcolor[HTML]{57BB8A}40.14 & \cellcolor[HTML]{64B2C0}40.67 & \cellcolor[HTML]{8CD1AF}37.29 & \cellcolor[HTML]{7CBECA}60.19 \\
% \midrule
% bloom-560m              & \cellcolor[HTML]{C0E6D3}36.14 & \cellcolor[HTML]{A7D4DC}50.11 & \cellcolor[HTML]{C2E7D5}35.37 & \cellcolor[HTML]{FAFDFD}-2.6  & \cellcolor[HTML]{BFE5D3}33.41 & \cellcolor[HTML]{BFE0E6}31.85 \\
% bloom-1b1               & \cellcolor[HTML]{D3EDE0}34.08 & \cellcolor[HTML]{7EBFCB}68.6  & \cellcolor[HTML]{B7E2CD}35.86 & \cellcolor[HTML]{EBF6F7}1.63  & \cellcolor[HTML]{A5DBC1}35.37 & \cellcolor[HTML]{98CCD5}48.52 \\
% bloom-3b                & \cellcolor[HTML]{B7E2CD}37.15 & \cellcolor[HTML]{76BBC8}72.28 & \cellcolor[HTML]{BFE6D3}35.48 & \cellcolor[HTML]{DCEEF1}5.94  & \cellcolor[HTML]{9FD8BC}35.85 & \cellcolor[HTML]{87C4CF}55.33 \\
% bloom-7b1               & \cellcolor[HTML]{B9E3CE}36.96 & \cellcolor[HTML]{69B5C3}77.82 & \cellcolor[HTML]{93D4B4}37.47 & \cellcolor[HTML]{D0E8EC}9.53  & \cellcolor[HTML]{9BD7BA}36.12 & \cellcolor[HTML]{7ABEC9}61    \\
 & \texttt{BLOOM}                   & \cellcolor[HTML]{A0D9BD}39.72 & \cellcolor[HTML]{6CB7C4}76.63 & \cellcolor[HTML]{8DD1B0}37.72 & \cellcolor[HTML]{C7E4E9}11.95 & \cellcolor[HTML]{89D0AD}37.48 & \cellcolor[HTML]{7ABDC9}61.17 \\
% bloomz-560m             & \cellcolor[HTML]{C9EADA}35.12 & \cellcolor[HTML]{BBDEE4}41.21 & \cellcolor[HTML]{F4FBF7}33.14 & \cellcolor[HTML]{FCFEFE}-3.08 & \cellcolor[HTML]{FFFFFF}28.55 & \cellcolor[HTML]{E1F1F3}17.56 \\
% bloomz-1b1              & \cellcolor[HTML]{CAEADA}35.00 & \cellcolor[HTML]{6CB7C4}76.66 & \cellcolor[HTML]{BCE4D0}35.65 & \cellcolor[HTML]{DBEDF1}6.4   & \cellcolor[HTML]{A7DCC2}35.22 & \cellcolor[HTML]{8AC6D0}54.19 \\
% bloomz-3b               & \cellcolor[HTML]{C4E7D6}35.74 & \cellcolor[HTML]{6EB8C5}75.86 & \cellcolor[HTML]{BBE4D0}35.68 & \cellcolor[HTML]{D3EAEE}8.56  & \cellcolor[HTML]{ADDEC6}34.75 & \cellcolor[HTML]{8EC8D2}52.51 \\
% bloomz-7b1              & \cellcolor[HTML]{B8E2CE}37.05 & \cellcolor[HTML]{67B4C2}79.09 & \cellcolor[HTML]{A2DABF}36.78 & \cellcolor[HTML]{D0E8EC}9.43  & \cellcolor[HTML]{9AD6B9}36.21 & \cellcolor[HTML]{7DBFCB}59.53 \\
 & \texttt{BLOOMZ}                  & \cellcolor[HTML]{B3E0CA}37.63 & \cellcolor[HTML]{60B1BF}82.06 & \cellcolor[HTML]{A6DBC2}36.6  & \cellcolor[HTML]{C4E2E7}12.9  & \cellcolor[HTML]{8FD2B1}37.06 & \cellcolor[HTML]{66B3C1}69.55 \\
% \midrule
% gpt-j-6b                & \cellcolor[HTML]{A7DCC2}38.86 & \cellcolor[HTML]{6CB7C4}76.48 & \cellcolor[HTML]{AFDFC8}36.2  & \cellcolor[HTML]{CCE6EB}10.53 & \cellcolor[HTML]{92D3B3}36.8  & \cellcolor[HTML]{7DBFCB}59.59 \\
% gpt-neox-20b            & \cellcolor[HTML]{A6DBC1}39.04 & \cellcolor[HTML]{6EB8C5}75.81 & \cellcolor[HTML]{B3E1CB}36.02 & \cellcolor[HTML]{CCE6EB}10.62 & \cellcolor[HTML]{91D3B3}36.87 & \cellcolor[HTML]{84C2CD}56.85 \\
% llama-7b                & \cellcolor[HTML]{97D5B7}40.70 & \cellcolor[HTML]{6FB8C5}75.52 & \cellcolor[HTML]{9FD8BC}36.95 & \cellcolor[HTML]{CDE7EB}10.28 & \cellcolor[HTML]{94D4B4}36.7  & \cellcolor[HTML]{88C4CF}55.31 \\
% llama-13b               & \cellcolor[HTML]{99D6B8}40.45 & \cellcolor[HTML]{6DB7C4}76.13 & \cellcolor[HTML]{9ED8BC}36.98 & \cellcolor[HTML]{C9E5EA}11.43 & \cellcolor[HTML]{8ED1B0}37.16 & \cellcolor[HTML]{7DBFCB}59.61 \\
% llama-30b               & \cellcolor[HTML]{A5DBC0}39.14 & \cellcolor[HTML]{67B4C2}78.74 & \cellcolor[HTML]{91D3B3}37.56 & \cellcolor[HTML]{C7E3E8}12.21 & \cellcolor[HTML]{89D0AD}37.5  & \cellcolor[HTML]{73BAC7}63.89 \\
% llama-65b               & \cellcolor[HTML]{AADDC4}38.59 & \cellcolor[HTML]{61B1BF}81.59 & \cellcolor[HTML]{8AD0AE}37.86 & \cellcolor[HTML]{C7E3E8}12.2  & \cellcolor[HTML]{7BCAA3}38.59 & \cellcolor[HTML]{6AB6C3}67.82 \\
 & \texttt{OPT-1.3b}                & \cellcolor[HTML]{DCF1E7}33.01 & \cellcolor[HTML]{6EB8C5}75.57 & \cellcolor[HTML]{E0F3EA}34    & \cellcolor[HTML]{E4F2F4}3.82  & \cellcolor[HTML]{ADDEC6}34.76 & \cellcolor[HTML]{92CAD3}50.78 \\
% opt-6.7b                & \cellcolor[HTML]{A9DDC4}38.64 & \cellcolor[HTML]{6CB7C4}76.62 & \cellcolor[HTML]{D0ECDF}34.73 & \cellcolor[HTML]{D2E9ED}8.86  & \cellcolor[HTML]{95D4B5}36.58 & \cellcolor[HTML]{7FC0CC}58.68 \\
% opt-13b                 & \cellcolor[HTML]{A8DCC3}38.78 & \cellcolor[HTML]{67B4C2}79.08 & \cellcolor[HTML]{D1EDDF}34.69 & \cellcolor[HTML]{D3E9ED}8.73  & \cellcolor[HTML]{87CFAB}37.67 & \cellcolor[HTML]{7BBECA}60.65 \\
 & \texttt{OPT-30b}                 & \cellcolor[HTML]{AFDFC7}38.04 & \cellcolor[HTML]{6BB6C3}77.22 & \cellcolor[HTML]{C8E9D9}35.08 & \cellcolor[HTML]{CEE7EC}9.96  & \cellcolor[HTML]{88CFAC}37.58 & \cellcolor[HTML]{78BDC9}61.79 \\
% opt-66b                 & \cellcolor[HTML]{A0D9BD}39.64 & \cellcolor[HTML]{6CB7C4}76.72 & \cellcolor[HTML]{BAE3CF}35.72 & \cellcolor[HTML]{C9E5EA}11.42 & \cellcolor[HTML]{8AD0AD}37.45 & \cellcolor[HTML]{7BBECA}60.43 \\
% \midrule
 & \texttt{OPT-IML-MAX-1.3b}        & \cellcolor[HTML]{C1E6D4}36.00 & \cellcolor[HTML]{65B3C1}79.73 & \cellcolor[HTML]{9DD8BB}37.01 & \cellcolor[HTML]{C8E4E9}11.85 & \cellcolor[HTML]{8FD2B1}37.08 & \cellcolor[HTML]{76BCC8}62.68 \\
 & \texttt{OPT-IML-MAX-30b}         & \cellcolor[HTML]{8BD0AE}42.03 & \cellcolor[HTML]{66B4C1}79.39 & \cellcolor[HTML]{B8E3CE}35.8  & \cellcolor[HTML]{C8E4E9}11.73 & \cellcolor[HTML]{6DC49A}39.59 & \cellcolor[HTML]{6DB7C4}66.39 \\
 & \texttt{Flan-T5-small}           & \cellcolor[HTML]{AADDC4}38.57 & \cellcolor[HTML]{6BB6C3}77.26 & \cellcolor[HTML]{A5DBC1}36.65 & \cellcolor[HTML]{6BB6C3}38.6  & \cellcolor[HTML]{86CEAB}37.72 & \cellcolor[HTML]{69B5C3}68.15 \\
 & \texttt{Flan-T5-base}            & \cellcolor[HTML]{90D3B2}41.40 & \cellcolor[HTML]{65B3C1}79.7  & \cellcolor[HTML]{A2DABF}36.79 & \cellcolor[HTML]{64B3C0}40.63 & \cellcolor[HTML]{7AC9A2}38.67 & \cellcolor[HTML]{69B5C3}68.09 \\
 & \texttt{Flan-T5-large}          & \cellcolor[HTML]{89D0AD}42.17 & \cellcolor[HTML]{64B3C0}80.44 & \cellcolor[HTML]{BAE4CF}35.71 & \cellcolor[HTML]{61B1BF}41.31 & \cellcolor[HTML]{74C79E}39.08 & \cellcolor[HTML]{64B3C1}70.27 \\
 & \texttt{Flan-T5-xl}              & \cellcolor[HTML]{93D4B4}41.07 & \cellcolor[HTML]{59ADBC}85.06 & \cellcolor[HTML]{F2FAF6}33.21 & \cellcolor[HTML]{57ACBB}44.12 & \cellcolor[HTML]{89CFAD}37.51 & \cellcolor[HTML]{57ACBB}75.5  \\
 & \texttt{Flan-T5-xxl}             & \cellcolor[HTML]{8DD1B0}41.75 & \cellcolor[HTML]{5BAEBD}84.13 & \cellcolor[HTML]{DAF1E6}34.27 & \cellcolor[HTML]{5AAEBC}43.43 & \cellcolor[HTML]{70C59B}39.42 & \cellcolor[HTML]{5DAFBE}73.05 \\
% ul2                     & \cellcolor[HTML]{C4E8D6}35.65 & \cellcolor[HTML]{C4E2E7}37.01 & \cellcolor[HTML]{B6E2CD}35.89 & \cellcolor[HTML]{B5DBE1}17.15 & \cellcolor[HTML]{A7DCC2}35.22 & \cellcolor[HTML]{B3DAE0}37.07 \\
% \midrule
 & \texttt{Flan-UL2}                & \cellcolor[HTML]{83CDA9}42.83 & \cellcolor[HTML]{5BAEBD}84.34 & \cellcolor[HTML]{C3E7D6}35.31 & \cellcolor[HTML]{5CAFBD}42.8  & \cellcolor[HTML]{64C193}40.27 & \cellcolor[HTML]{5DAFBE}73.23 \\
\bottomrule
\end{tabular}
}
\caption{For brevity, we report automatic metrics for simplification (SARI) and meaning preservation (BERTScore) for select models using Prompt 2. `\texttt{*}' indicates closed-weights. The full list of results is available in Tables \ref{table:asset-results}, \ref{table:med-easi-results}, and \ref{table:newsela-results} in the Appendix.}
\label{table:results-summary} 
\end{table*}

% Briefly compare prompt performance (already mentioned above)
% \paragraph{Prompts} 
\paragraph{Structured prompting improves performance.}
Figure \ref{fig:all_prompts_violin} reveals that prompts 0 and 2 both offer a slight advantage over prompt 1, especially in regard to meaning preservation. 
This confirms that providing a structured template for few-shot examples instead of embedding them within sentences is the most beneficial. Hence, we focus on prompt 2 for all our analysis, as it provides the most detailed description of the task and has also been used in prior work \cite{maddela-etal-2023-lens}. 
% In terms of SARI, the distribution of scores under prompt 2 is wider, indicating that while the highly detailed instructions benefit some models, others struggle to leverage this information. 
% As a result, we find prompt 1 performs most consistently with all models tested. 
% Nevertheless, we choose to focus on results for prompt 2 in the remaining analysis as this prompt is the most for the task.
% which outperforms prompt 0 by a small margin, provides the most detailed description of the task and target audience . 

% % Talk about model size
% \paragraph{Model Size} 
% To visualize model performance, we plot SARI and BERTScores in \ref{fig:sari_bert_boxplots} for a selected set of LLMs.
% Models on the x-axis are ordered by model family, and within each model family, they are ordered by size (ascending). 
% The general and expected pattern is that as model size increases, performance improves. 
% This pattern is most visible for SARI, albeit with notable exceptions. For example, \texttt{Flan-T5-large} has higher SARI on \asset than its two larger models \texttt{Flan-T5-xl} (3 billion parameters) and \texttt{Flan-T5-xxl} (11 billion parameters).

% \paragraph{It's not the size, but the way you train it.} 
\paragraph{Training method matters more than size.}
Table \ref{table:results-summary} presents the performance according to SARI and BERTScore for the top-performing LLMs.
Scaling LLMs has revealed strong benefits in few-shot settings \citep{brown2020language, chowdhery2022palm}; however, in our evaluation, we observe numerous exceptions to this rule. For example, \texttt{Flan-T5-large} (770 million parameters) consistently attains higher SARI scores on \asset than \texttt{Flan-T5-xl} (3 billion parameters) and \texttt{Flan-T5-xxl} (11 billion parameters).\footnote{We include a wider comparison of selected LLMs on \asset in Figure \ref{fig:sari_bert_boxplots} in the Appendix.}
Meanwhile, we observe that training strategies such as instruction-tuning and RLHF help to deliver greater improvements, especially for meaning preservation, as measured by BERTScore.
% For example, \texttt{T5-lm-adapt} models are typically among the worst performers in our few-shot setting. 
This agrees with previous findings that demonstrate the benefits of instruction-based adaption strategies for improved generalization abilities \citep{schick-schutze-2021-just, zhang_differentiable_2022, chung2022scaling}. 

% Additionally, we summarize the main findings for each dataset according to SARI, which rewards correct n-gram edit operations, and BERTScore, which rewards meaning preservation. Among the 3 test datasets, models generally perform the best on ASSET, second best on Newsela-Manual, and the worst on Med-EASi. It makes sense that the medical domain is the most difficult TS domain. We believe that since most of these LLMs have likely been exposed to ASSET's domain of Wikipedia during training, it could have influenced its performance on ASSET.

\paragraph{\asset}

% sari and bertscore

On Wikipedia-style data, OpenAI’s \texttt{Davinci-003} and \texttt{GPT-3.5-Turbo} outperform all other tested LLMs by a considerable margin according to SARI.
Strikingly, these models also outperform the ground truth references, which are closely approximated by the previous state-of-the-art \texttt{MUSS} models. 
This is notable since \texttt{MUSS-wiki-mined} was trained on the in-domain \ts dataset of WikiLarge \citep{zhang-lapata-2017-sentence}.
% \paragraph{BERTScore} When considering meaning preservation as represented by BERTScore, we again observe strong performances from OpenAI’s models, with \texttt{davinci-002} outperforming all others on \asset. 
Meanwhile, for open-weight contenders, we can see in Table \ref{table:results-summary} that only a small number of models are competitive, namely \texttt{OPT-IML-Max-30b}, \texttt{Flan-T5-large}, and \texttt{Flan-UL2}, which scores the best balance between simplicity and meaning preservation according to automatic metrics.
% the open-weight alternative of \texttt{Flan-T5-xl} performs competitively on \asset.

\paragraph{\medeasi}

For medical-related texts, we observe that the majority of the models consistently fail to preserve meaning (our qualitative analysis in Section \ref{sec:qualitative_analysis} confirms this, see Table \ref{table:human-eval-results-all}).
The drop in meaning preservation can likely be explained by the fact that models are known to produce inadequate generations in out-of-domain settings \citep{muller-etal-2020-domain, singhal-etal-2023-assessing}. 
The models that do strike a reasonable balance with both SARI and BERTScore are again OpenAI's more powerful offerings and the Flan models. Notably, we also observe that the two MUSS models are able to perform competitively with the Flan models despite being multiple orders of magnitude smaller. 
% (see Tables \ref{table:med-easi-results} and \ref{table:newsela-results}).

\paragraph{\newsela}

Evaluating LLMs on professionally written simplifications from \newsela reveals that even the best LLMs are not able to match human performance.
This is observable through the clear margins of around 20 SARI points and 14 BERTScore points between the best performers and the gold simplifications.
On this dataset, \texttt{MUSS-wiki-mined} remains a strong baseline, outperforming all LLMs on both metrics, while \texttt{Davinci-002}, \texttt{Flan-UL2}, and  \texttt{Flan-T5-xxl} show the strongest performances among the LLMs.

\begin{figure}[t!]
    \centering
    \includegraphics[width=\linewidth]{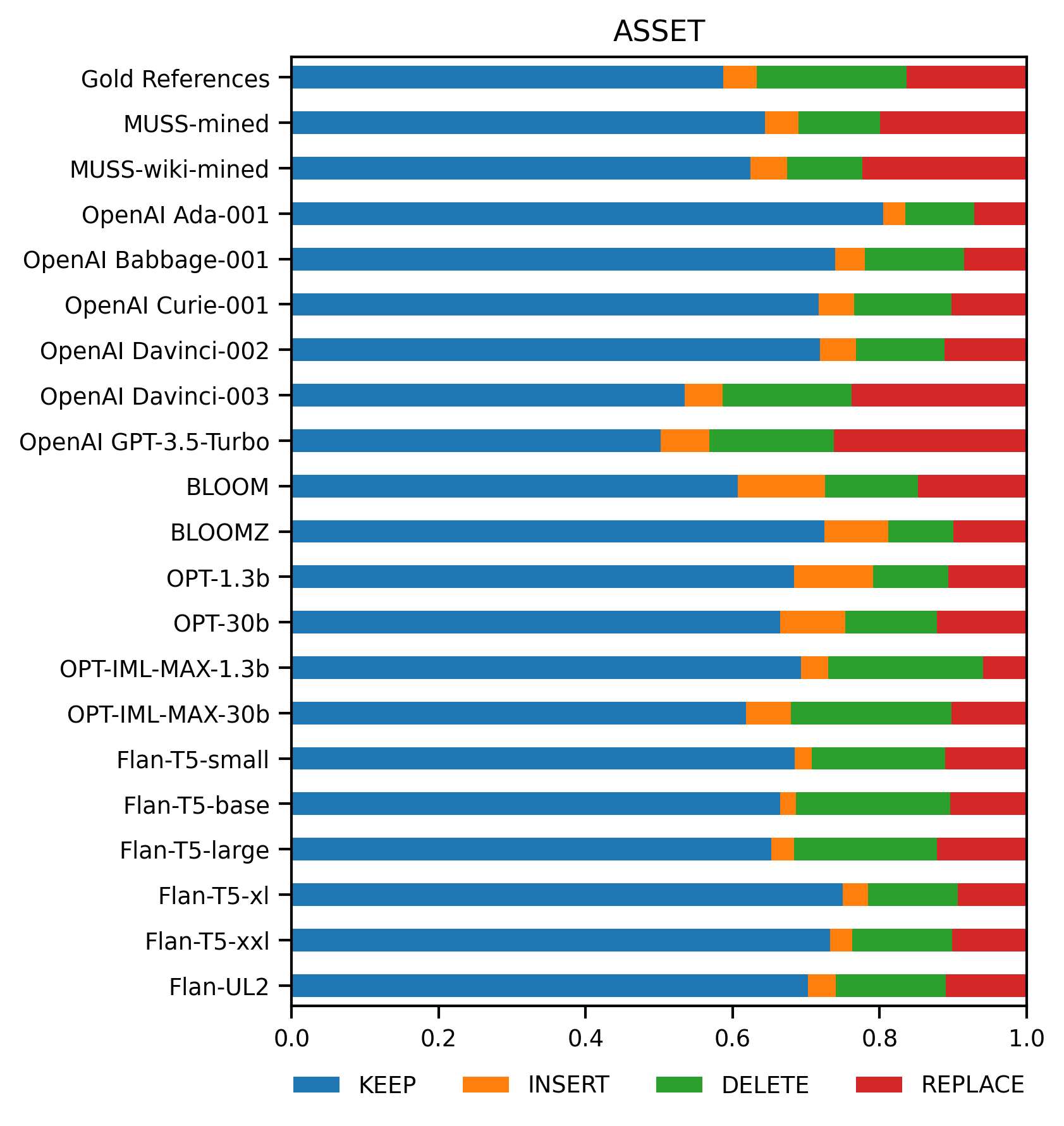}
    \caption{Distribution of token-level edit operations produced by the best-performing LLMs.}
    \label{fig:edit_ops_subset}
\end{figure}

\subsection{Analysis of Edit Operations}
\label{sec:quantitative_analysis}

% We quantify the simplification operations employed by the LLMs. 
To identify the main token-level edit operations performed by LLMs, we use an adaptation of the Wagner–Fischer algorithm \citep{Wagner1974}, following previous work by \citet{vasquez-rodriguez-etal-2021-investigating}. Specifically, we calculate the portion of insertion, replacement, deletion, and keep operations between the input source sentence and each of the system outputs for each dataset.
% changes between the predictions and references with respect to
% to determine changes at the token level between each complex and predicted sentence
% For the automatic annotation of simplification operations, we use the implementation from \citet{vasquez-rodriguez-etal-2021-investigating}.\footnote{\url{https://github.com/lmvasque/ts-explore}}

% \input{tables/annotation_scheme.tex}

Figure \ref{fig:edit_ops_subset} shows the distribution of token-level edit operations for the best-performing LLMs on \asset (for a more comprehensive view across all datasets and models, see Figure \ref{fig:edit_ops} in the Appendix).
Most models perform all four operations to differing degrees; however, similar to the gold references, the keep operation is by far the most prominent in this dataset.
Notably, \texttt{Davinci-003} and \texttt{GPT-3.5-Turbo} perform the most diverse set of operations, with fewer additions and more replacements than other models.
Insertions are typically less frequent, suggesting that the majority of the models avoid adding new and potentially irrelevant content.
% While deletions are often an easy way to simplify a piece of text, it can also result the loss of potentially important information and is therefore less desirable. 
We observe that most LLMs are within the range of the gold references in terms of the amount of information they delete when simplifying.
% eplacement operations appear to be most challenging for the LLMs, with only \texttt{MUSS}, \texttt{davinci-003} and \texttt{GPT-3.5-Turbo} exceeding or even matching the gold references.

% MUSS baselines and the smaller Flan-T5 models approximate the edit operations performed in the gold references most closely. 
% Meanwhile, majority of the models rely heavily on KEEP and INSERT operations, while the OpenAI models have more instances of the REPLACE operation but still a low number of deletions, which is most inline with the gold references. Strikingly, LLMs consistently over-rely on the KEEP operation, indicating minimal rewriting.

\begin{table*}[thb]
\centering
\footnotesize
% \scalebox{0.5}{}{%
\begin{tabular}{@{}l|cccccccc@{}}
\toprule
\textbf{Model outputs} 
  & \textbf{\%S$\uparrow$} &
  \textbf{\%MP$\uparrow$} &
  \textbf{\%L+} &
  \textbf{\%P+} & 
  \textbf{\%D+} & 
  \textbf{\%Sp+} & 
  \textbf{\%R+} &
  \textbf{\%H$\downarrow$} \\
\midrule \midrule
All  & 61.67  & 67.33 & 30.33 & 28.33 & 35.0 & 4.33 & 4.67 & 12.33 \\
\midrule 
% cohere-command-light & 39.05 & 7.85 & 68.02 & 52 \\
Top 5 SARI & 72.0 & 68.0 & \textbf{48.0} & \textbf{34.66} & 37.33 & \textbf{6.67} & 6.67 & 8.0 \\
Top 5 BERT & 62.67 & \textbf{84.0} & 17.33 & 29.33 & 34.67 & 5.33 & \textbf{10.67} & \textbf{2.67} \\
Top 5 FKGL & 34.67 & 40.0 & 14.66 & 17.33 & 26.67 & 0.0 & 0.0 & 36.0 \\
Top 5 LENS & \textbf{77.33} & 77.33 & 41.33 & 32.0 & \textbf{41.33} & 5.33 & 1.33 & \textbf{2.67} \\
\midrule
Open-Weight & 58.58 & 64.55 & 29.47 & 22.76  & \textbf{36.94} & 3.36 & 3.73 & 13.81 \\
Closed-Weight & \textbf{87.50} & \textbf{90.63} & \textbf{37.50} & \textbf{75.0} & 18.75 & \textbf{12.50} & \textbf{12.50} & \textbf{0.0} \\
\midrule
On \asset  & \textbf{77.0}  & \textbf{82.0} & 31.0 & \textbf{54.0} & 33.0 & \textbf{8.0} & 4.0 & \textbf{10.0} \\
On \newsela & 54.0 & 70.0 & \textbf{34.0} & 9.0 & \textbf{38.0} & 5.0 & \textbf{7.0} & 17.0 \\
On \medeasi & 54.0 & 50.0 & 26.0 & 22.0 & 34.0 & 0.0 & 3.0 & \textbf{10.0} \\
\bottomrule
\end{tabular}
% }
\caption{Results of our manual analysis.
The annotation schema includes the following annotation features:
S↑: accepted simplification,
% G$\downarrow$: ungrammatical, 
MP↑: meaning preserved, 
L+: lexical simplification, 
P+: paraphrasing,
R+: reordering (no changes),
D+: deletion, 
Sp+: sentence splitting, 
H$\downarrow$: hallucination.
% NS: no simplification.
% We exclude the Grammaticality column because all were nearly 100\% and the No Simplification column because the \%S column accounts for this. 
% We focus our analysis on simplicity (SARI) and meaning preservation (BERTScore).
}
\label{table:human-eval-results-all} 

\end{table*}

\section{Qualitative Analysis}
\label{sec:qualitative_analysis}
Automatic metrics are known to have blind spots and are not always entirely reliable \citep{alva-manchego-etal-2021-un, he_blind_2022}.
To compensate for this, we perform a qualitative analysis on a total of 300 system outputs. 

% \subsection{Setup}
First, we check whether or not each output is a valid simplification and highlight common failure cases such as inappropriate changes to the meaning of the original text, ungrammatical outputs, and the occurrence of hallucinations. 
Then, we annotate occurrences of common simplification edit operations such as lexical simplification, deletion, sentence splitting, reordering, and paraphrasing.\footnote{All annotations were completed by one of the authors and validated separately by another.}
% Table \ref{tab:annotation_scheme} shows an overview of our annotation scheme.
% \footnote{We refrain from evaluating system outputs using rating scales (e.g. Likert) for ranking simplifications, as they are more subjective and less informative for our case study.}  

For our annotations, we select model outputs from the top five systems ranked according to performance on the individual evaluation metrics of SARI, BERTScore, FKGL, and LENS. 
In each ranking set, we randomly select five complex-simple pairs from all generation settings.
% For each dataset, we consider outputs generated with all three prompts used to perform ICL.
To evaluate a unique set of models for greater diversity, if a system is repeated in the ranking (e.g.\ two different prompt types from the same model appear in the top five), we choose the next best system for analysis. 
An example of our annotated outputs is shown in Table \ref{tab:annotation_scheme_sari_example} in the Appendix. 
Table \ref{table:human-eval-results-all} shows results from this analysis, which we describe according to different criteria below.
% We omit top 5 FKGL and LENS model results for brevity and because they provide less insight.\footnote{FKGL can be gamed using degenerations \citep{tanprasert-kauchak-2021-flesch}, which we observe in our annotations. As a learned metric, LENS is not easily interpretable.}
% We define good performance as having outputs that contain a diverse set of simplification operations without impacting meaning preservation and that is both grammatical and adequate. 
\paragraph{By Automatic Metric}
Overall, we find that simplicity and meaning preservation are fairly balanced. 
% However, simplicity is traded off with meaning preservation if we only consider the top 5 models for individual dimensions. 
However, there is a clear trade-off between these two axes when we consider the top 5 models according to SARI and BERTScore. 
This agrees with earlier findings from \citet{Schwarzer2018}.
Along with a higher degree of simplicity, the top 5 SARI models exhibit more diverse edit operations than those ranked highly by BERTScore.
% But the top 5 SARI models simplify outputs more than the top 5 BERT models, while the opposite is true for meaning preservation. This supports our previous assertion that SARI indicates simplification level and BERTScore indicates the level of meaning preservation.

LENS, however, does not trade off simplicity and meaning preservation and even achieves a higher simplicity score than SARI along with its increased level of deletion. This result is in line with the previous finding that LENS achieves stronger correlations with human judgments compared to existing \ts metrics \citep{maddela-etal-2023-lens}. 
The top 5 models ranked by FKGL, on the other hand, produce outputs with low simplicity and meaning preservation and an especially high amount of hallucinations. This result supports the previous finding that FKGL can be easily gamed by degenerations \citep{tanprasert-kauchak-2021-flesch} and is therefore an unsuitable metric for evaluating the outputs of automatic \ts systems.

\paragraph{By Open-Status}
Open-weight models most frequently use the operations of lexical simplification, paraphrasing, and deletion, while structural operations such as sentence splitting and reordering are often neglected. Many only achieve high meaning preservation by directly copying the input sentence. 
However, the closed-weight models investigated here behave very differently: they produce close to 10\% more splitting, lexical simplification, and re-ordering than open-weight ones, while simultaneously performing fewer deletions. 
This leads to a greater degree of paraphrasing.
% The fact that they change the output so much while maintaining the highest levels of both simplicity and meaning preservation and never hallucinating is very impressive.
% Finally, closed-weight models outperform open-weight ones by a large margin.

\begin{figure*}[h]
\centering
\includegraphics[width=\linewidth]{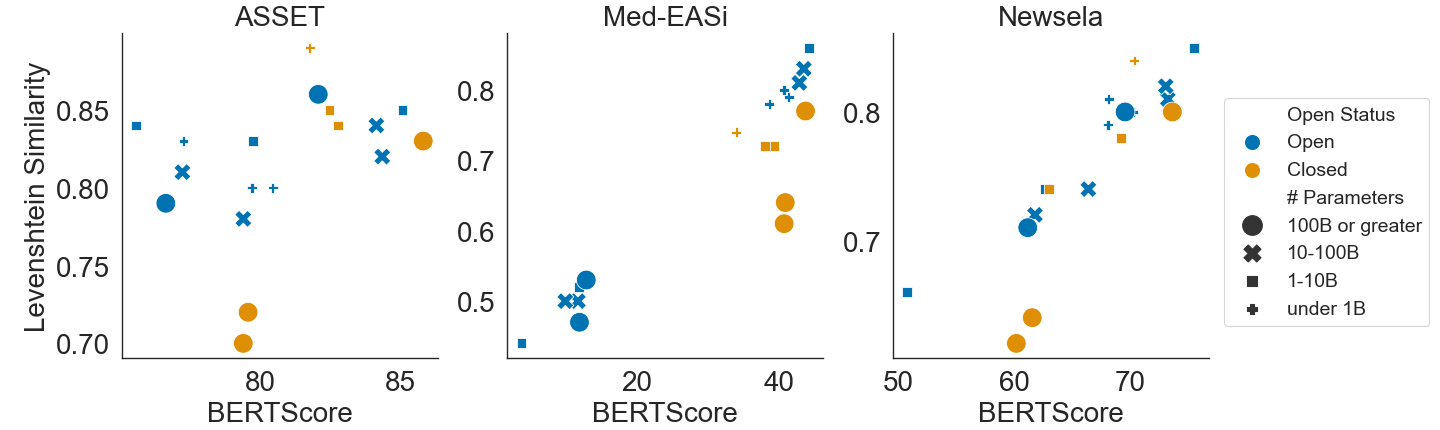}
\caption{BERTScore, computed between the system output and reference sentence(s), correlates strongly with Levenshtein similarity, computed between the source sentence and system outputs. 
This indicates that BERTScore tends to reward minimally edited sentences. 
Levenshtein similarity is computed with the EASSE package \citep{alva-manchego-etal-2019-easse}.}
\label{fig:bert_vs_lev_sim}
\end{figure*}

\paragraph{By Domain}
When comparing performance between different domains, we observe that all LLMs do significantly better on general encyclopedic texts in \asset in terms of both simplicity and meaning preservation, while also exhibiting a diverse set of edit operations. Although outputs from \newsela contain more hallucinations, meaning preservation is still fairly high. Outputs from \medeasi, on the other hand, have the lowest meaning preservation by far and the least diverse set of edit operations.
We find that \medeasi outputs, along with others that do not preserve meaning, often contain repetitions, hallucinations, and in some cases even copy the input prompt, demonstrating a tendency to disregard the instruction and thus fail to complete the task. 
% Generally, we find that the OpenAI models perform consistently well across annotation metrics and domains,
These failure modes are most frequently observed from the smaller \texttt{T5} models, but are also exhibited by models such as \texttt{LLaMA} when evaluated on \medeasi. 

\section{Discussion}

We discuss our results around the following aspects: the access level of the simplification models (open- vs.~closed-weight), the training strategies (general pre-training vs.~general fine-tuning strategies), and the utility of automatic metrics.

% Architecture / Access level / type: 
% Discussion the performance of open vs closed models
\paragraph{Access Level} 
Among the OpenAI models, we observe that all models perform particularly well on meaning preservation according to BERTScore but exhibit considerable differences in their ability to simplify, as indicated by SARI on `weaker' models such as \texttt{Ada-001}. 
Among the evaluated open-weight models, we observe that the Flan models (\texttt{T5} and \texttt{UL2}) typically perform competitively, punching well above their weight in terms of parameter counts with much larger decoder-only models.
This is a promising finding for the category of open-weight models, and we hope that this encourages future work to continue investigating different methods regardless of the model size.

% General pretraining, instruction tuning & Reinforcement Learning from Human Feedback (RLHF).
\paragraph{Training Strategies} 

Within model families, when comparing base models to their instruction fine-tuned counterparts, we observe that instruction-tuning typically leads to better performance in our few-shot ICL setting for \ts.
% This is in line with the findings from \citet{chung2022scaling} that suggest instruction fine-tuning serves as a general-purpose method to improve pre-trained LMs.
We find this to be particularly encouraging since \ts is one task often hindered by the lack of high-quality labeled training data \citep{stajner-2021-automatic}.

% Comparing model families that share the same base model but differ in their alignment strategy (e.g.~instruction fine-tuning), we observe that instruction-tuned models often perform better than their unaligned counterparts.
Nevertheless, improvement is not always guaranteed, as seen when comparing \texttt{BLOOM} vs \texttt{BLOOMZ}. In this case, instruction fine-tuning leads to better-meaning preservation but a reduction in the degree of simplification, indicating that the instruction-tuning method used to derive the multilingual \texttt{BLOOMZ} may be less suitable for English \ts. This stands in stark contrast to the Flan instruction tuning method, which delivers considerable gains in both SARI and BERTScore despite sharing the same underlying instruction-tuning dataset as \texttt{BLOOMZ}.
Therefore, we hypothesize that this drop in performance may be influenced by the multilingual instruction tuning setup that is unique to \texttt{BLOOMZ}.

% results for all - BERT score
\paragraph{Utility of Automatic Metrics} 
Overall, we find SARI and BERTScore to be useful automatic evaluation metrics for inspecting the trade-off between simplicity and meaning preservation (see Figure \ref{fig:sari_vs_bert} in the Appendix). 
% shows the relationship between these two metrics for models grouped by status and size.
% While we do not observe any specific patterns according to these groupings, we can see that for the different domains,
% BERTScls groupedore generally increases with the number of parameters, indicating that larger models are typically better at retaining the meaning of the original sentence. 
% However, this contrasts with the degree of simplification according to SARI, where larger models often fail to match that of some of their smaller siblings. 
% Closed-weight models often strike a more optimal balance given this trade-off. 
In general, closed-weight models often strike a more optimal balance. This is also supported by our qualitative analysis, which confirmed that these models rely less on deletion, an oft-overused operation \cite{devaraj-etal-2022-evaluating}, and more on other edits (e.g. paraphrasing or splitting).
% results for 3 datasets - SARI vs BERT
 % Content deletion is one of the most prominent operations in text simplification . While this can increase the simplicity of the output as shown by SARI, it can also compromise the meaning conveyed in the original text. 
% When the text is simplified, it loses meaning as the content will be likely to be deleted or replaced. 
% This highlights an important trade-off between meaning preservation and the degree of simplification that needs to be considered when evaluating TS.

Furthermore, our qualitative analysis shows that outputs with higher BERTScores tend to be minimally simplified, often copying the entire input text.
% For BERTScore, we noticed that the best models showed that system outputs that were not simplified at all or with minimal changes at a lexical level. 
We validate this by studying the relationship between BERTScore (computed between the system output and the reference sentence(s)) and Levenshtein similarity (computed between the system output and the original input sentence).  
Figure~\ref{fig:bert_vs_lev_sim} reveals a strong positive correlation across all datasets, indicating that BERTScore tends to reward minimally simplified responses.
For some of the closed-models, which tend to perform a greater degree of paraphrasing, this leads to lower BERTScores, while models that perform more copying are rewarded. 
% highlight a limitation of the references in the datasets. suggests that reference sentences in the evaluation datasets remain similar to the
Overall, the results from our qualitative study generally showed agreement with those from our automatic evaluation metrics, particularly SARI, BERTScore, and LENS. It also enabled us to pinpoint specific operations, such as re-ordering, and identify issues, notably hallucinations, in system outputs.

\section{Conclusion}
In this paper, we provided a comprehensive assessment of how well out-of-the-box LLMs perform on the task of \ts with few-shot in-context learning. We found that the best LLMs outperform state-of-the-art supervised \ts baselines while also producing a more diverse set of simplification operations. We also established that closed-weight models perform better than open-weight ones and that general instruction-tuning often improves a model's abilities on \ts. Furthermore, we empirically validated the trade-off between simplicity and meaning preservation through automatic evaluation and a manual analysis. 
Our analyses of multiple few-shot prompting strategies revealed that a more structured prompting format produces better results than presenting source-target examples in continuous text.

Our performance benchmark, BLESS, provides a strong foundation for future work. For example, it remains an open question as to which expressions and instructions are optimal for prompting LLMs to simplify texts. Furthermore, this work exclusively focused on few-shot in-context learning. Future work could explore the capabilities of these systems in zero-shot, fine-tuned, or retrieval-based settings.

\section*{Limitations}

In this section, we discuss a few limitations of our work. 
First, we only considered English \ts datasets, and it still remains to be seen how these \ts abilities transfer to languages other than English. 
Additionally, we selected only a handful of output samples for manual analysis for the three test datasets considered, and 
all annotations were performed by one of the authors and subsequently validated by another author independently. 
It will be necessary to perform this at a larger scale to more accurately characterize the capabilities of each model for each domain and prompt. We further acknowledge the limits of the evaluation set itself. While we purposefully chose the test splits to cover a variety of domains, test splits for all three corpora amount to 915 samples, which could potentially limit the statistical power of results obtained from the assessment. 
Additionally, two out of the three test sets contain only sentences as input, while the third contains short multi-sentence texts, so this assessment mostly applies to the subtask of sentence simplification.
Finally, our findings confirm that proprietary, closed-source models can achieve a new state-of-the-art performance on the task of text simplification. 
However, very little is known about their training data, alignment strategies, and implementation behind paywalled APIs.
Therefore, the comparison to open-source models, which contain no explicit training on the task and an extremely bare-bones implementation is potentially unfair.
% further acknowledge that these test sets may potentially be used in pre-training stages of newly released models, which can heavily skew the obtained results.

% EMNLP 2023 requires all submissions to have a section titled ``Limitations'', for discussing the limitations of the paper as a complement to the discussion of strengths in the main text. This section should occur after the conclusion, but before the references. It will not count towards the page limit.  

% The discussion of limitations is mandatory. Papers without a limitation section will be desk-rejected without review.
% ARR-reviewed papers that did not include ``Limitations'' section in their prior submission, should submit a PDF with such a section together with their EMNLP 2023 submission.

% While we are open to different types of limitations, just mentioning that a set of results have been shown for English only probably does not reflect what we expect. 
% Mentioning that the method works mostly for languages with limited morphology, like English, is a much better alternative.
% In addition, limitations such as low scalability to long text, the requirement of large GPU resources, or other things that inspire crucial further investigation are welcome.

\section*{Ethics Statement}

This work is conducted in full awareness of and in line with the ACL Ethics Policy. Particularly, this work contributes to the transparency and fairness of evaluation methodologies in line with Sections 1.1, 1.2, 2.7, and 2.9 of the code, which innately leads to avoiding seen and unseen harms (Section 1.2, 1.4). We contribute to improving expertise in the domain of text simplification (Section 2.6). All models, datasets, and compute resources are used with permission and with concern to the appropriate access rights and licenses (Section 2.8). Our work contributes to the professional development of the research team (Section 3.5) and more widely benefits the research community and wider society (Section 3.1) by augmenting the understanding of the capacity of LLMs on the specific task of \ts. 

% Scientific work published at EMNLP 2023 must comply with the \href{https://www.aclweb.org/portal/content/acl-code-ethics}{ACL Ethics Policy}. We encourage all authors to include an explicit ethics statement on the broader impact of the work, or other ethical considerations after the conclusion but before the references. The ethics statement will not count toward the page limit (8 pages for long, 4 pages for short papers).

\section*{Acknowledgements}

We would like to thank Sian Gooding for her initiative in motivating this project, as well as Hoang Nguyen Hung Van, Jan Trienes, and everyone in the text simplification research community who joined our discussions during this journey. 
Thank you also to the anonymous reviewers for providing valuable feedback.
This work was facilitated by the infrastructure services provided by S3IT, the Service and Support for Science IT team at the University of Zurich.
Laura Vásquez-Rodríguez’s work was funded by the Kilburn Scholarship from the University of Manchester.

% \bibliography{anthology,zotero_tk, custom}
\bibliography{anthology, custom}

\begin{thebibliography}{78}
\expandafter\ifx\csname natexlab\endcsname\relax\def\natexlab#1{#1}\fi

\bibitem[{Agrawal et~al.(2023)Agrawal, Zhou, Lewis, Zettlemoyer, and
  Ghazvininejad}]{agrawal-etal-2023-context}
Sweta Agrawal, Chunting Zhou, Mike Lewis, Luke Zettlemoyer, and Marjan
  Ghazvininejad. 2023.
\newblock \href {https://doi.org/10.18653/v1/2023.findings-acl.564} {In-context
  examples selection for machine translation}.
\newblock In \emph{Findings of the Association for Computational Linguistics:
  ACL 2023}, pages 8857--8873, Toronto, Canada. Association for Computational
  Linguistics.

\bibitem[{Alva-Manchego et~al.(2020{\natexlab{a}})Alva-Manchego, Martin,
  Bordes, Scarton, Sagot, and Specia}]{alva-manchego-etal-2020-asset}
Fernando Alva-Manchego, Louis Martin, Antoine Bordes, Carolina Scarton,
  Beno{\^\i}t Sagot, and Lucia Specia. 2020{\natexlab{a}}.
\newblock \href {https://doi.org/10.18653/v1/2020.acl-main.424} {{ASSET}: {A}
  dataset for tuning and evaluation of sentence simplification models with
  multiple rewriting transformations}.
\newblock In \emph{Proceedings of the 58th Annual Meeting of the Association
  for Computational Linguistics}, pages 4668--4679, Online. Association for
  Computational Linguistics.

\bibitem[{Alva-Manchego et~al.(2019{\natexlab{a}})Alva-Manchego, Martin,
  Scarton, and Specia}]{alva-manchego-etal-2019-easse}
Fernando Alva-Manchego, Louis Martin, Carolina Scarton, and Lucia Specia.
  2019{\natexlab{a}}.
\newblock \href {https://doi.org/10.18653/v1/D19-3009} {{EASSE}: Easier
  automatic sentence simplification evaluation}.
\newblock In \emph{Proceedings of the 2019 Conference on Empirical Methods in
  Natural Language Processing and the 9th International Joint Conference on
  Natural Language Processing (EMNLP-IJCNLP): System Demonstrations}, pages
  49--54, Hong Kong, China. Association for Computational Linguistics.

\bibitem[{Alva-Manchego et~al.(2019{\natexlab{b}})Alva-Manchego, Scarton, and
  Specia}]{alva-manchego-etal-2019-cross}
Fernando Alva-Manchego, Carolina Scarton, and Lucia Specia. 2019{\natexlab{b}}.
\newblock \href {https://aclanthology.org/W19-3656} {Cross-sentence
  transformations in text simplification}.
\newblock In \emph{Proceedings of the 2019 Workshop on Widening NLP}, pages
  181--184, Florence, Italy. Association for Computational Linguistics.

\bibitem[{Alva-Manchego et~al.(2020{\natexlab{b}})Alva-Manchego, Scarton, and
  Specia}]{alva-manchego-etal-2020-data}
Fernando Alva-Manchego, Carolina Scarton, and Lucia Specia. 2020{\natexlab{b}}.
\newblock \href {https://doi.org/10.1162/coli_a_00370} {Data-driven sentence
  simplification: Survey and benchmark}.
\newblock \emph{Computational Linguistics}, 46(1):135--187.

\bibitem[{Alva-Manchego et~al.(2021)Alva-Manchego, Scarton, and
  Specia}]{alva-manchego-etal-2021-un}
Fernando Alva-Manchego, Carolina Scarton, and Lucia Specia. 2021.
\newblock \href {https://doi.org/10.1162/coli_a_00418} {The (un)suitability of
  automatic evaluation metrics for text simplification}.
\newblock \emph{Computational Linguistics}, 47(4):861--889.

\bibitem[{Aumiller and Gertz(2022)}]{aumiller-gertz-2022-unihd}
Dennis Aumiller and Michael Gertz. 2022.
\newblock \href {https://doi.org/10.18653/v1/2022.tsar-1.28} {{U}ni{HD} at
  {TSAR}-2022 shared task: Is compute all we need for lexical simplification?}
\newblock In \emph{Proceedings of the Workshop on Text Simplification,
  Accessibility, and Readability (TSAR-2022)}, pages 251--258, Abu Dhabi,
  United Arab Emirates (Virtual). Association for Computational Linguistics.

\bibitem[{Baidoo-Anu and Owusu~Ansah(2023)}]{baidoo2023education}
David Baidoo-Anu and Leticia Owusu~Ansah. 2023.
\newblock Education in the era of generative artificial intelligence (ai):
  Understanding the potential benefits of chatgpt in promoting teaching and
  learning.
\newblock \emph{Available at SSRN 4337484}.

\bibitem[{Basu et~al.(2023)Basu, Vasu, Yasunaga, and Yang}]{basu2023medeasi}
Chandrayee Basu, Rosni Vasu, Michihiro Yasunaga, and Qian Yang. 2023.
\newblock \href {https://doi.org/10.1609/aaai.v37i12.26649} {Med-easi: Finely
  annotated dataset and models for controllable simplification of medical
  texts}.
\newblock In \emph{Proceedings of the Thirty-Seventh AAAI Conference on
  Artificial Intelligence and Thirty-Fifth Conference on Innovative
  Applications of Artificial Intelligence and Thirteenth Symposium on
  Educational Advances in Artificial Intelligence}, AAAI'23/IAAI'23/EAAI'23.
  AAAI Press.

\bibitem[{Baumgartner et~al.(2020)Baumgartner, Zannettou, Keegan, Squire, and
  Blackburn}]{baumgartner2020pushshift}
Jason Baumgartner, Savvas Zannettou, Brian Keegan, Megan Squire, and Jeremy
  Blackburn. 2020.
\newblock The pushshift reddit dataset.
\newblock In \emph{Proceedings of the international AAAI conference on web and
  social media}, volume~14, pages 830--839.

\bibitem[{Black et~al.(2022)Black, Biderman, Hallahan, Anthony, Gao, Golding,
  He, Leahy, McDonell, Phang, Pieler, Prashanth, Purohit, Reynolds, Tow, Wang,
  and Weinbach}]{black-etal-2022-gpt}
Sidney Black, Stella Biderman, Eric Hallahan, Quentin Anthony, Leo Gao,
  Laurence Golding, Horace He, Connor Leahy, Kyle McDonell, Jason Phang,
  Michael Pieler, Usvsn~Sai Prashanth, Shivanshu Purohit, Laria Reynolds,
  Jonathan Tow, Ben Wang, and Samuel Weinbach. 2022.
\newblock \href {https://doi.org/10.18653/v1/2022.bigscience-1.9}
  {{GPT}-{N}eo{X}-20{B}: An open-source autoregressive language model}.
\newblock In \emph{Proceedings of BigScience Episode {\#}5 -- Workshop on
  Challenges {\&} Perspectives in Creating Large Language Models}, pages
  95--136, virtual+Dublin. Association for Computational Linguistics.

\bibitem[{Brown et~al.(2020)Brown, Mann, Ryder, Subbiah, Kaplan, Dhariwal,
  Neelakantan, Shyam, Sastry, Askell et~al.}]{brown2020language}
Tom Brown, Benjamin Mann, Nick Ryder, Melanie Subbiah, Jared~D Kaplan, Prafulla
  Dhariwal, Arvind Neelakantan, Pranav Shyam, Girish Sastry, Amanda Askell,
  et~al. 2020.
\newblock Language models are few-shot learners.
\newblock \emph{Advances in neural information processing systems},
  33:1877--1901.

\bibitem[{Chi et~al.(2023)Chi, Chen, Chang, Lee, and Chang}]{chi2023learning}
Alison Chi, Li-Kuang Chen, Yi-Chen Chang, Shu-Hui Lee, and Jason~S. Chang.
  2023.
\newblock \href {http://arxiv.org/abs/2308.02226} {{Learning to Paraphrase
  Sentences to Different Complexity Levels}}.
\newblock \emph{arXiv preprint arXiv:2308.02226}.

\bibitem[{Chowdhery et~al.(2022)Chowdhery, Narang, Devlin, Bosma, Mishra,
  Roberts, Barham, Chung, Sutton, Gehrmann et~al.}]{chowdhery2022palm}
Aakanksha Chowdhery, Sharan Narang, Jacob Devlin, Maarten Bosma, Gaurav Mishra,
  Adam Roberts, Paul Barham, Hyung~Won Chung, Charles Sutton, Sebastian
  Gehrmann, et~al. 2022.
\newblock Palm: Scaling language modeling with pathways.
\newblock \emph{arXiv preprint arXiv:2204.02311}.

\bibitem[{Chung et~al.(2022)Chung, Hou, Longpre, Zoph, Tay, Fedus, Li, Wang,
  Dehghani, Brahma et~al.}]{chung2022scaling}
Hyung~Won Chung, Le~Hou, Shayne Longpre, Barret Zoph, Yi~Tay, William Fedus,
  Eric Li, Xuezhi Wang, Mostafa Dehghani, Siddhartha Brahma, et~al. 2022.
\newblock Scaling instruction-finetuned language models.
\newblock \emph{arXiv preprint arXiv:2210.11416}.

\bibitem[{Dettmers et~al.(2022)Dettmers, Lewis, Belkada, and
  Zettlemoyer}]{dettmers_llmint8_2022}
Tim Dettmers, Mike Lewis, Younes Belkada, and Luke Zettlemoyer. 2022.
\newblock \href {http://arxiv.org/abs/2208.07339} {{LLM}.int8(): 8-bit {Matrix}
  {Multiplication} for {Transformers} at {Scale}}.
\newblock ArXiv:2208.07339 [cs].

\bibitem[{Devaraj et~al.(2022)Devaraj, Sheffield, Wallace, and
  Li}]{devaraj-etal-2022-evaluating}
Ashwin Devaraj, William Sheffield, Byron Wallace, and Junyi~Jessy Li. 2022.
\newblock \href {https://doi.org/10.18653/v1/2022.acl-long.506} {Evaluating
  factuality in text simplification}.
\newblock In \emph{Proceedings of the 60th Annual Meeting of the Association
  for Computational Linguistics (Volume 1: Long Papers)}, pages 7331--7345,
  Dublin, Ireland. Association for Computational Linguistics.

\bibitem[{Dowling and Lucey(2023)}]{Dowling_2023}
Michael Dowling and Brian Lucey. 2023.
\newblock \href {https://doi.org/https://doi.org/10.1016/j.frl.2023.103662}
  {Chatgpt for (finance) research: The bananarama conjecture}.
\newblock \emph{Finance Research Letters}, 53:103662.

\bibitem[{Feng et~al.(2023)Feng, Qiang, Li, Yuan, and Zhu}]{feng_sentence_2023}
Yutao Feng, Jipeng Qiang, Yun Li, Yunhao Yuan, and Yi~Zhu. 2023.
\newblock \href {http://arxiv.org/abs/2302.11957} {Sentence {Simplification}
  via {Large} {Language} {Models}}.
\newblock ArXiv:2302.11957 [cs].

\bibitem[{Gao et~al.(2020)Gao, Biderman, Black, Golding, Hoppe, Foster, Phang,
  He, Thite, Nabeshima et~al.}]{gao2020pile}
Leo Gao, Stella Biderman, Sid Black, Laurence Golding, Travis Hoppe, Charles
  Foster, Jason Phang, Horace He, Anish Thite, Noa Nabeshima, et~al. 2020.
\newblock The pile: An 800gb dataset of diverse text for language modeling.
\newblock \emph{arXiv preprint arXiv:2101.00027}.

\bibitem[{He et~al.(2022)He, Zhang, Wang, Kumar, Cho, Glass, and
  Tsvetkov}]{he_blind_2022}
Tianxing He, Jingyu Zhang, Tianle Wang, Sachin Kumar, Kyunghyun Cho, James
  Glass, and Yulia Tsvetkov. 2022.
\newblock \href {http://arxiv.org/abs/2212.10020} {On the {Blind} {Spots} of
  {Model}-{Based} {Evaluation} {Metrics} for {Text} {Generation}}.
\newblock ArXiv:2212.10020 [cs].

\bibitem[{Holtzman et~al.(2020)Holtzman, Buys, Du, Forbes, and
  Choi}]{holtzman_curious_2020}
Ari Holtzman, Jan Buys, Li~Du, Maxwell Forbes, and Yejin Choi. 2020.
\newblock \href {http://arxiv.org/abs/1904.09751} {The {Curious} {Case} of
  {Neural} {Text} {Degeneration}}.
\newblock \emph{arXiv:1904.09751 [cs]}.

\bibitem[{Iyer et~al.(2023)Iyer, Lin, Pasunuru, Mihaylov, Simig, Yu, Shuster,
  Wang, Liu, Koura, Li, O'Horo, Pereyra, Wang, Dewan, Celikyilmaz, Zettlemoyer,
  and Stoyanov}]{iyer2023optiml}
Srinivasan Iyer, Xi~Victoria Lin, Ramakanth Pasunuru, Todor Mihaylov, Daniel
  Simig, Ping Yu, Kurt Shuster, Tianlu Wang, Qing Liu, Punit~Singh Koura, Xian
  Li, Brian O'Horo, Gabriel Pereyra, Jeff Wang, Christopher Dewan, Asli
  Celikyilmaz, Luke Zettlemoyer, and Ves Stoyanov. 2023.
\newblock \href {http://arxiv.org/abs/2212.12017} {Opt-iml: Scaling language
  model instruction meta learning through the lens of generalization}.

\bibitem[{Jiang et~al.(2020)Jiang, Maddela, Lan, Zhong, and
  Xu}]{jiang-etal-2020-neural}
Chao Jiang, Mounica Maddela, Wuwei Lan, Yang Zhong, and Wei Xu. 2020.
\newblock \href {https://doi.org/10.18653/v1/2020.acl-main.709} {Neural {CRF}
  model for sentence alignment in text simplification}.
\newblock In \emph{Proceedings of the 58th Annual Meeting of the Association
  for Computational Linguistics}, pages 7943--7960, Online. Association for
  Computational Linguistics.

\bibitem[{Kincaid et~al.(1975)Kincaid, Fishburne, Rogers, and
  Chissom}]{Kincaid-1975}
J.~Peter Kincaid, Robert~P. Fishburne, R~L Rogers, and Brad~S. Chissom. 1975.
\newblock \href {https://stars.library.ucf.edu/istlibrary/56/} {Derivation of
  new readability formulas (automated readability index, fog count and flesch
  reading ease formula) for navy enlisted personnel}.
\newblock In \emph{Institute for Simulation and Training}, pages 1--49.

\bibitem[{Lauren{\c{c}}on et~al.(2022)Lauren{\c{c}}on, Saulnier, Wang, Akiki,
  Villanova~del Moral, Le~Scao, Von~Werra, Mou, Gonz{\'a}lez~Ponferrada, Nguyen
  et~al.}]{laurenccon2022bigscience}
Hugo Lauren{\c{c}}on, Lucile Saulnier, Thomas Wang, Christopher Akiki, Albert
  Villanova~del Moral, Teven Le~Scao, Leandro Von~Werra, Chenghao Mou, Eduardo
  Gonz{\'a}lez~Ponferrada, Huu Nguyen, et~al. 2022.
\newblock The bigscience roots corpus: A 1.6 tb composite multilingual dataset.
\newblock \emph{Advances in Neural Information Processing Systems},
  35:31809--31826.

\bibitem[{Lester et~al.(2021)Lester, Al-Rfou, and
  Constant}]{lester-etal-2021-power}
Brian Lester, Rami Al-Rfou, and Noah Constant. 2021.
\newblock \href {https://doi.org/10.18653/v1/2021.emnlp-main.243} {The power of
  scale for parameter-efficient prompt tuning}.
\newblock In \emph{Proceedings of the 2021 Conference on Empirical Methods in
  Natural Language Processing}, pages 3045--3059, Online and Punta Cana,
  Dominican Republic. Association for Computational Linguistics.

\bibitem[{Lewis et~al.(2020)Lewis, Liu, Goyal, Ghazvininejad, Mohamed, Levy,
  Stoyanov, and Zettlemoyer}]{lewis-etal-2020-bart}
Mike Lewis, Yinhan Liu, Naman Goyal, Marjan Ghazvininejad, Abdelrahman Mohamed,
  Omer Levy, Veselin Stoyanov, and Luke Zettlemoyer. 2020.
\newblock \href {https://doi.org/10.18653/v1/2020.acl-main.703} {{BART}:
  Denoising sequence-to-sequence pre-training for natural language generation,
  translation, and comprehension}.
\newblock In \emph{Proceedings of the 58th Annual Meeting of the Association
  for Computational Linguistics}, pages 7871--7880, Online. Association for
  Computational Linguistics.

\bibitem[{Lhoest et~al.(2021)Lhoest, del Moral, Jernite, Thakur, von Platen,
  Patil, Chaumond, Drame, Plu, Tunstall et~al.}]{lhoest2021datasets}
Quentin Lhoest, Albert~Villanova del Moral, Yacine Jernite, Abhishek Thakur,
  Patrick von Platen, Suraj Patil, Julien Chaumond, Mariama Drame, Julien Plu,
  Lewis Tunstall, et~al. 2021.
\newblock Datasets: A community library for natural language processing.
\newblock \emph{arXiv preprint arXiv:2109.02846}.

\bibitem[{Li et~al.(2022)Li, Zhao, Lyu, and Wang}]{li-etal-2022-eliciting}
Yanyang Li, Jianqiao Zhao, Michael Lyu, and Liwei Wang. 2022.
\newblock \href {https://doi.org/10.18653/v1/2022.emnlp-main.721} {Eliciting
  knowledge from large pre-trained models for unsupervised knowledge-grounded
  conversation}.
\newblock In \emph{Proceedings of the 2022 Conference on Empirical Methods in
  Natural Language Processing}, pages 10551--10564, Abu Dhabi, United Arab
  Emirates. Association for Computational Linguistics.

\bibitem[{Liu et~al.(2023)Liu, Iter, Xu, Wang, Xu, and Zhu}]{liu2023geval}
Yang Liu, Dan Iter, Yichong Xu, Shuohang Wang, Ruochen Xu, and Chenguang Zhu.
  2023.
\newblock \href
  {https://www.microsoft.com/en-us/research/publication/gpteval-nlg-evaluation-using-gpt-4-with-better-human-alignment/}
  {G-eval: Nlg evaluation using gpt-4 with better human alignment}.
\newblock \emph{arXiv 2303.16634}.

\bibitem[{Lu et~al.(2022)Lu, Bartolo, Moore, Riedel, and
  Stenetorp}]{lu_fantastically_2022}
Yao Lu, Max Bartolo, Alastair Moore, Sebastian Riedel, and Pontus Stenetorp.
  2022.
\newblock \href {https://doi.org/10.18653/v1/2022.acl-long.556} {Fantastically
  {Ordered} {Prompts} and {Where} to {Find} {Them}: {Overcoming} {Few}-{Shot}
  {Prompt} {Order} {Sensitivity}}.
\newblock In \emph{Proceedings of the 60th {Annual} {Meeting} of the
  {Association} for {Computational} {Linguistics} ({Volume} 1: {Long}
  {Papers})}, pages 8086--8098, Dublin, Ireland. Association for Computational
  Linguistics.

\bibitem[{Maddela et~al.(2023)Maddela, Dou, Heineman, and
  Xu}]{maddela-etal-2023-lens}
Mounica Maddela, Yao Dou, David Heineman, and Wei Xu. 2023.
\newblock \href {https://doi.org/10.18653/v1/2023.acl-long.905} {{LENS}: A
  learnable evaluation metric for text simplification}.
\newblock In \emph{Proceedings of the 61st Annual Meeting of the Association
  for Computational Linguistics (Volume 1: Long Papers)}, pages 16383--16408,
  Toronto, Canada. Association for Computational Linguistics.

\bibitem[{Martin et~al.(2020)Martin, de~la Clergerie, Sagot, and
  Bordes}]{martin-etal-2020-controllable}
Louis Martin, {\'E}ric de~la Clergerie, Beno{\^\i}t Sagot, and Antoine Bordes.
  2020.
\newblock \href {https://aclanthology.org/2020.lrec-1.577} {Controllable
  sentence simplification}.
\newblock In \emph{Proceedings of the Twelfth Language Resources and Evaluation
  Conference}, pages 4689--4698, Marseille, France. European Language Resources
  Association.

\bibitem[{Martin et~al.(2022)Martin, Fan, de~la Clergerie, Bordes, and
  Sagot}]{martin-etal-2022-muss}
Louis Martin, Angela Fan, {\'E}ric de~la Clergerie, Antoine Bordes, and
  Beno{\^\i}t Sagot. 2022.
\newblock \href {https://aclanthology.org/2022.lrec-1.176} {{MUSS}:
  Multilingual unsupervised sentence simplification by mining paraphrases}.
\newblock In \emph{Proceedings of the Thirteenth Language Resources and
  Evaluation Conference}, pages 1651--1664, Marseille, France. European
  Language Resources Association.

\bibitem[{Muennighoff et~al.(2022)Muennighoff, Wang, Sutawika, Roberts,
  Biderman, Scao, Bari, Shen, Yong, Schoelkopf
  et~al.}]{muennighoff2022crosslingual}
Niklas Muennighoff, Thomas Wang, Lintang Sutawika, Adam Roberts, Stella
  Biderman, Teven~Le Scao, M~Saiful Bari, Sheng Shen, Zheng-Xin Yong, Hailey
  Schoelkopf, et~al. 2022.
\newblock Crosslingual generalization through multitask finetuning.
\newblock \emph{arXiv preprint arXiv:2211.01786}.

\bibitem[{M{\"u}ller et~al.(2020)M{\"u}ller, Rios, and
  Sennrich}]{muller-etal-2020-domain}
Mathias M{\"u}ller, Annette Rios, and Rico Sennrich. 2020.
\newblock \href {https://aclanthology.org/2020.amta-research.14} {Domain
  robustness in neural machine translation}.
\newblock In \emph{Proceedings of the 14th Conference of the Association for
  Machine Translation in the Americas (Volume 1: Research Track)}, pages
  151--164, Virtual. Association for Machine Translation in the Americas.

\bibitem[{Ouyang et~al.(2022)Ouyang, Wu, Jiang, Almeida, Wainwright, Mishkin,
  Zhang, Agarwal, Slama, Ray, Schulman, Hilton, Kelton, Miller, Simens, Askell,
  Welinder, Christiano, Leike, and Lowe}]{ouyang2022training}
Long Ouyang, Jeff Wu, Xu~Jiang, Diogo Almeida, Carroll~L. Wainwright, Pamela
  Mishkin, Chong Zhang, Sandhini Agarwal, Katarina Slama, Alex Ray, John
  Schulman, Jacob Hilton, Fraser Kelton, Luke Miller, Maddie Simens, Amanda
  Askell, Peter Welinder, Paul Christiano, Jan Leike, and Ryan Lowe. 2022.
\newblock \href {http://arxiv.org/abs/2203.02155} {Training language models to
  follow instructions with human feedback}.

\bibitem[{Paetzold and Specia(2016)}]{paetzold-specia-2016-benchmarking}
Gustavo Paetzold and Lucia Specia. 2016.
\newblock \href {https://aclanthology.org/L16-1491} {Benchmarking lexical
  simplification systems}.
\newblock In \emph{Proceedings of the Tenth International Conference on
  Language Resources and Evaluation ({LREC}'16)}, pages 3074--3080,
  Portoro{\v{z}}, Slovenia. European Language Resources Association (ELRA).

\bibitem[{Papineni et~al.(2002)Papineni, Roukos, Ward, and
  Zhu}]{papineni-etal-2002-bleu}
Kishore Papineni, Salim Roukos, Todd Ward, and Wei-Jing Zhu. 2002.
\newblock \href {https://doi.org/10.3115/1073083.1073135} {{B}leu: a method for
  automatic evaluation of machine translation}.
\newblock In \emph{Proceedings of the 40th Annual Meeting of the Association
  for Computational Linguistics}, pages 311--318, Philadelphia, Pennsylvania,
  USA. Association for Computational Linguistics.

\bibitem[{Radford et~al.(2018)Radford, Narasimhan, Salimans, Sutskever
  et~al.}]{radford2018improving}
Alec Radford, Karthik Narasimhan, Tim Salimans, Ilya Sutskever, et~al. 2018.
\newblock Improving language understanding by generative pre-training.
\newblock Technical report, OpenAI.

\bibitem[{Radford et~al.(2019)Radford, Wu, Child, Luan, Amodei, Sutskever, and
  {others}}]{radford2019language}
Alec Radford, Jeffrey Wu, Rewon Child, David Luan, Dario Amodei, Ilya
  Sutskever, and {others}. 2019.
\newblock Language models are unsupervised multitask learners.
\newblock \emph{OpenAI blog}, 1(8):9.

\bibitem[{Raffel et~al.(2020)Raffel, Shazeer, Roberts, Lee, Narang, Matena,
  Zhou, Li, and Liu}]{raffel2020exploring}
Colin Raffel, Noam Shazeer, Adam Roberts, Katherine Lee, Sharan Narang, Michael
  Matena, Yanqi Zhou, Wei Li, and Peter~J. Liu. 2020.
\newblock \href {http://jmlr.org/papers/v21/20-074.html} {Exploring the limits
  of transfer learning with a unified text-to-text transformer}.
\newblock \emph{Journal of Machine Learning Research}, 21(140):1--67.

\bibitem[{Ryan et~al.(2023)Ryan, Naous, and Xu}]{ryan-etal-2023-revisiting}
Michael Ryan, Tarek Naous, and Wei Xu. 2023.
\newblock \href {https://doi.org/10.18653/v1/2023.acl-long.269} {Revisiting
  non-{E}nglish text simplification: A unified multilingual benchmark}.
\newblock In \emph{Proceedings of the 61st Annual Meeting of the Association
  for Computational Linguistics (Volume 1: Long Papers)}, pages 4898--4927,
  Toronto, Canada. Association for Computational Linguistics.

\bibitem[{Saggion et~al.(2022)Saggion, {\v{S}}tajner, Ferr{\'e}s, Sheang,
  Shardlow, North, and Zampieri}]{saggion-etal-2022-findings}
Horacio Saggion, Sanja {\v{S}}tajner, Daniel Ferr{\'e}s, Kim~Cheng Sheang,
  Matthew Shardlow, Kai North, and Marcos Zampieri. 2022.
\newblock \href {https://doi.org/10.18653/v1/2022.tsar-1.31} {Findings of the
  {TSAR}-2022 shared task on multilingual lexical simplification}.
\newblock In \emph{Proceedings of the Workshop on Text Simplification,
  Accessibility, and Readability (TSAR-2022)}, pages 271--283, Abu Dhabi,
  United Arab Emirates (Virtual). Association for Computational Linguistics.

\bibitem[{Sallam(2023)}]{Malik_2023}
Malik Sallam. 2023.
\newblock \href {https://doi.org/10.3390/healthcare11060887} {Chatgpt utility
  in healthcare education, research, and practice: Systematic review on the
  promising perspectives and valid concerns}.
\newblock \emph{Healthcare}, 11(6).

\bibitem[{Sanh et~al.(2022)Sanh, Webson, Raffel, Bach, Sutawika, Alyafeai,
  Chaffin, Stiegler, Raja, Dey, Bari, Xu, Thakker, Sharma, Szczechla, Kim,
  Chhablani, Nayak, Datta, Chang, Jiang, Wang, Manica, Shen, Yong, Pandey,
  Bawden, Wang, Neeraj, Rozen, Sharma, Santilli, F{\'{e}}vry, Fries, Teehan,
  Scao, Biderman, Gao, Wolf, and Rush}]{sanh2022multitask}
Victor Sanh, Albert Webson, Colin Raffel, Stephen~H. Bach, Lintang Sutawika,
  Zaid Alyafeai, Antoine Chaffin, Arnaud Stiegler, Arun Raja, Manan Dey,
  M~Saiful Bari, Canwen Xu, Urmish Thakker, Shanya~Sharma Sharma, Eliza
  Szczechla, Taewoon Kim, Gunjan Chhablani, Nihal~V. Nayak, Debajyoti Datta,
  Jonathan Chang, Mike~Tian{-}Jian Jiang, Han Wang, Matteo Manica, Sheng Shen,
  Zheng~Xin Yong, Harshit Pandey, Rachel Bawden, Thomas Wang, Trishala Neeraj,
  Jos Rozen, Abheesht Sharma, Andrea Santilli, Thibault F{\'{e}}vry, Jason~Alan
  Fries, Ryan Teehan, Teven~Le Scao, Stella Biderman, Leo Gao, Thomas Wolf, and
  Alexander~M. Rush. 2022.
\newblock \href {https://openreview.net/forum?id=9Vrb9D0WI4} {Multitask
  prompted training enables zero-shot task generalization}.
\newblock In \emph{The Tenth International Conference on Learning
  Representations, {ICLR} 2022, Virtual Event, April 25-29, 2022}.
  OpenReview.net.

\bibitem[{Scao et~al.(2022)Scao, Fan, Akiki, Pavlick, Ili{\'c}, Hesslow,
  Castagn{\'e}, Luccioni, Yvon, Gall{\'e} et~al.}]{scao2022bloom}
Teven~Le Scao, Angela Fan, Christopher Akiki, Ellie Pavlick, Suzana Ili{\'c},
  Daniel Hesslow, Roman Castagn{\'e}, Alexandra~Sasha Luccioni, Fran{\c{c}}ois
  Yvon, Matthias Gall{\'e}, et~al. 2022.
\newblock Bloom: A 176b-parameter open-access multilingual language model.
\newblock \emph{arXiv preprint arXiv:2211.05100}.

\bibitem[{Schick and Sch{\"u}tze(2021)}]{schick-schutze-2021-just}
Timo Schick and Hinrich Sch{\"u}tze. 2021.
\newblock \href {https://doi.org/10.18653/v1/2021.naacl-main.185} {It{'}s not
  just size that matters: Small language models are also few-shot learners}.
\newblock In \emph{Proceedings of the 2021 Conference of the North American
  Chapter of the Association for Computational Linguistics: Human Language
  Technologies}, pages 2339--2352, Online. Association for Computational
  Linguistics.

\bibitem[{Schwarzer and Kauchak(2018)}]{Schwarzer2018}
Max Schwarzer and David Kauchak. 2018.
\newblock {Human Evaluation for Text Simplification: The Simplicity-Adequacy
  Tradeoff}.
\newblock Technical report, SoCal NLP Symposium.

\bibitem[{Singhal et~al.(2023)Singhal, Forristal, Ye, and
  Durrett}]{singhal-etal-2023-assessing}
Prasann Singhal, Jarad Forristal, Xi~Ye, and Greg Durrett. 2023.
\newblock \href {https://doi.org/10.18653/v1/2023.eacl-main.175} {Assessing
  out-of-domain language model performance from few examples}.
\newblock In \emph{Proceedings of the 17th Conference of the European Chapter
  of the Association for Computational Linguistics}, pages 2385--2397,
  Dubrovnik, Croatia. Association for Computational Linguistics.

\bibitem[{Sobania et~al.(2023)Sobania, Briesch, Hanna, and
  Petke}]{sobania2023analysis}
Dominik Sobania, Martin Briesch, Caril Hanna, and Justyna Petke. 2023.
\newblock \href {https://doi.org/10.1109/APR59189.2023.00012} {An analysis of
  the automatic bug fixing performance of chatgpt}.
\newblock In \emph{2023 IEEE/ACM International Workshop on Automated Program
  Repair (APR)}, pages 23--30, Los Alamitos, CA, USA. IEEE Computer Society.

\bibitem[{Stajner(2021)}]{stajner-2021-automatic}
Sanja Stajner. 2021.
\newblock \href {https://doi.org/10.18653/v1/2021.findings-acl.233} {Automatic
  text simplification for social good: Progress and challenges}.
\newblock In \emph{Findings of the Association for Computational Linguistics:
  ACL-IJCNLP 2021}, pages 2637--2652, Online. Association for Computational
  Linguistics.

\bibitem[{Stajner et~al.(2022)Stajner, Ferrés, Shardlow, North, Zampieri, and
  Saggion}]{Sanja_2022}
Sanja Stajner, Daniel Ferrés, Matthew Shardlow, Kai North, Marcos Zampieri,
  and Horacio Saggion. 2022.
\newblock \href {https://doi.org/10.3389/frai.2022.991242} {{Lexical
  simplification benchmarks for English, Portuguese, and Spanish}}.
\newblock \emph{Frontiers in Artificial Intelligence}, 5.

\bibitem[{Stiennon et~al.(2020)Stiennon, Ouyang, Wu, Ziegler, Lowe, Voss,
  Radford, Amodei, and Christiano}]{stiennon_learning_2020}
Nisan Stiennon, Long Ouyang, Jeffrey Wu, Daniel Ziegler, Ryan Lowe, Chelsea
  Voss, Alec Radford, Dario Amodei, and Paul~F Christiano. 2020.
\newblock \href
  {https://proceedings.neurips.cc/paper_files/paper/2020/file/1f89885d556929e98d3ef9b86448f951-Paper.pdf}
  {Learning to summarize with human feedback}.
\newblock In \emph{Advances in neural information processing systems},
  volume~33, pages 3008--3021. Curran Associates, Inc.

\bibitem[{Sun et~al.(2023)Sun, Xu, and Wan}]{sun-etal-2023-teaching}
Renliang Sun, Wei Xu, and Xiaojun Wan. 2023.
\newblock \href {https://doi.org/10.18653/v1/2023.findings-acl.595} {Teaching
  the pre-trained model to generate simple texts for text simplification}.
\newblock In \emph{Findings of the Association for Computational Linguistics:
  ACL 2023}, pages 9345--9355, Toronto, Canada. Association for Computational
  Linguistics.

\bibitem[{Tanprasert and Kauchak(2021)}]{tanprasert-kauchak-2021-flesch}
Teerapaun Tanprasert and David Kauchak. 2021.
\newblock \href {https://doi.org/10.18653/v1/2021.gem-1.1} {Flesch-kincaid is
  not a text simplification evaluation metric}.
\newblock In \emph{Proceedings of the 1st Workshop on Natural Language
  Generation, Evaluation, and Metrics (GEM 2021)}, pages 1--14, Online.
  Association for Computational Linguistics.

\bibitem[{Tay et~al.(2023)Tay, Dehghani, Tran, Garcia, Wei, Wang, Chung, Bahri,
  Schuster, Zheng, Zhou, Houlsby, and Metzler}]{tay2023ul}
Yi~Tay, Mostafa Dehghani, Vinh~Q. Tran, Xavier Garcia, Jason Wei, Xuezhi Wang,
  Hyung~Won Chung, Dara Bahri, Tal Schuster, Steven Zheng, Denny Zhou, Neil
  Houlsby, and Donald Metzler. 2023.
\newblock \href {https://openreview.net/forum?id=6ruVLB727MC} {{UL}2: Unifying
  language learning paradigms}.
\newblock In \emph{The Eleventh International Conference on Learning
  Representations}.

\bibitem[{Touvron et~al.(2023)Touvron, Lavril, Izacard, Martinet, Lachaux,
  Lacroix, Rozi{\`e}re, Goyal, Hambro, Azhar et~al.}]{touvron2023llama}
Hugo Touvron, Thibaut Lavril, Gautier Izacard, Xavier Martinet, Marie-Anne
  Lachaux, Timoth{\'e}e Lacroix, Baptiste Rozi{\`e}re, Naman Goyal, Eric
  Hambro, Faisal Azhar, et~al. 2023.
\newblock Llama: Open and efficient foundation language models.
\newblock \emph{arXiv preprint arXiv:2302.13971}.

\bibitem[{V{\'a}squez-Rodr{\'\i}guez et~al.(2022)V{\'a}squez-Rodr{\'\i}guez,
  Nguyen, Shardlow, and Ananiadou}]{vasquez-rodriguez-etal-2022-uom}
Laura V{\'a}squez-Rodr{\'\i}guez, Nhung Nguyen, Matthew Shardlow, and Sophia
  Ananiadou. 2022.
\newblock \href {https://doi.org/10.18653/v1/2022.tsar-1.23} {{U}o{M}{\&}{MMU}
  at {TSAR}-2022 shared task: Prompt learning for lexical simplification}.
\newblock In \emph{Proceedings of the Workshop on Text Simplification,
  Accessibility, and Readability (TSAR-2022)}, pages 218--224, Abu Dhabi,
  United Arab Emirates (Virtual). Association for Computational Linguistics.

\bibitem[{V{\'a}squez-Rodr{\'\i}guez
  et~al.(2021{\natexlab{a}})V{\'a}squez-Rodr{\'\i}guez, Shardlow, Przyby{\l}a,
  and Ananiadou}]{vasquez-rodriguez-etal-2021-investigating}
Laura V{\'a}squez-Rodr{\'\i}guez, Matthew Shardlow, Piotr Przyby{\l}a, and
  Sophia Ananiadou. 2021{\natexlab{a}}.
\newblock \href {https://doi.org/10.18653/v1/2021.findings-acl.77}
  {Investigating text simplification evaluation}.
\newblock In \emph{Findings of the Association for Computational Linguistics:
  ACL-IJCNLP 2021}, pages 876--882, Online. Association for Computational
  Linguistics.

\bibitem[{V{\'a}squez-Rodr{\'\i}guez
  et~al.(2021{\natexlab{b}})V{\'a}squez-Rodr{\'\i}guez, Shardlow, Przybyła,
  and Ananiadou}]{Vasquez-Rodriguez-2021b}
Laura V{\'a}squez-Rodr{\'\i}guez, Matthew Shardlow, Piotr Przybyła, and Sophia
  Ananiadou. 2021{\natexlab{b}}.
\newblock \href {http://ceur-ws.org/Vol-2944/paper4.pdf} {The role of text
  simplification operations in evaluation}.
\newblock In \emph{Proceedings of the First Workshop on Current Trends in Text
  Simplification (CTTS-2021)}, pages 57--69.

\bibitem[{Vaswani et~al.(2017)Vaswani, Shazeer, Parmar, Uszkoreit, Jones,
  Gomez, Kaiser, and Polosukhin}]{vaswani2017}
Ashish Vaswani, Noam Shazeer, Niki Parmar, Jakob Uszkoreit, Llion Jones,
  Aidan~N Gomez, \L~ukasz Kaiser, and Illia Polosukhin. 2017.
\newblock \href
  {https://proceedings.neurips.cc/paper_files/paper/2017/file/3f5ee243547dee91fbd053c1c4a845aa-Paper.pdf}
  {Attention is all you need}.
\newblock In \emph{Advances in Neural Information Processing Systems},
  volume~30. Curran Associates, Inc.

\bibitem[{Wagner and Fischer(1974)}]{Wagner1974}
Robert~A. Wagner and Michael~J. Fischer. 1974.
\newblock \href {https://doi.org/10.1145/321796.321811} {{The String-to-String
  Correction Problem}}.
\newblock \emph{Journal of the ACM (JACM)}, 21(1):168--173.

\bibitem[{Wang and Komatsuzaki(2021)}]{gpt-j}
Ben Wang and Aran Komatsuzaki. 2021.
\newblock {GPT-J-6B: A 6 Billion Parameter Autoregressive Language Model}.
\newblock \url{https://github.com/kingoflolz/mesh-transformer-jax}.

\bibitem[{Wang et~al.(2023)Wang, Lyu, Ji, Zhang, Yu, Shi, and
  Tu}]{wang2023documentlevel}
Longyue Wang, Chenyang Lyu, Tianbo Ji, Zhirui Zhang, Dian Yu, Shuming Shi, and
  Zhaopeng Tu. 2023.
\newblock \href {http://arxiv.org/abs/2304.02210} {Document-level machine
  translation with large language models}.

\bibitem[{Wang et~al.(2022)Wang, Mishra, Alipoormolabashi, Kordi, Mirzaei,
  Arunkumar, Ashok, Dhanasekaran, Naik, Stap, Pathak, Karamanolakis, Lai,
  Purohit, Mondal, Anderson, Kuznia, Doshi, Patel, Pal, Moradshahi, Parmar,
  Purohit, Varshney, Kaza, Verma, Puri, Karia, Sampat, Doshi, Mishra, Reddy,
  Patro, Dixit, Shen, Baral, Choi, Hajishirzi, Smith, and
  Khashabi}]{wang2022benchmarking}
Yizhong Wang, Swaroop Mishra, Pegah Alipoormolabashi, Yeganeh Kordi, Amirreza
  Mirzaei, Anjana Arunkumar, Arjun Ashok, Arut~Selvan Dhanasekaran, Atharva
  Naik, David Stap, Eshaan Pathak, Giannis Karamanolakis, Haizhi~Gary Lai,
  Ishan Purohit, Ishani Mondal, Jacob Anderson, Kirby Kuznia, Krima Doshi,
  Maitreya Patel, Kuntal~Kumar Pal, M.~Moradshahi, Mihir Parmar, Mirali
  Purohit, Neeraj Varshney, Phani~Rohitha Kaza, Pulkit Verma, Ravsehaj~Singh
  Puri, Rushang Karia, Shailaja~Keyur Sampat, Savan Doshi, Siddharth~Deepak
  Mishra, Sujan Reddy, Sumanta Patro, Tanay Dixit, Xudong Shen, Chitta Baral,
  Yejin Choi, Hannaneh Hajishirzi, Noah~A. Smith, and Daniel Khashabi. 2022.
\newblock \href {https://api.semanticscholar.org/CorpusID:248227391}
  {Benchmarking generalization via in-context instructions on 1, 600+ language
  tasks}.
\newblock \emph{ArXiv}, abs/2204.07705.

\bibitem[{Wei et~al.(2022)Wei, Bosma, Zhao, Guu, Yu, Lester, Du, Dai, and
  Le}]{wei2022finetuned}
Jason Wei, Maarten Bosma, Vincent Zhao, Kelvin Guu, Adams~Wei Yu, Brian Lester,
  Nan Du, Andrew~M. Dai, and Quoc~V Le. 2022.
\newblock \href {https://openreview.net/forum?id=gEZrGCozdqR} {Finetuned
  language models are zero-shot learners}.
\newblock In \emph{International Conference on Learning Representations}.

\bibitem[{Wolf et~al.(2020)Wolf, Debut, Sanh, Chaumond, Delangue, Moi, Cistac,
  Rault, Louf, Funtowicz, Davison, Shleifer, von Platen, Ma, Jernite, Plu, Xu,
  Le~Scao, Gugger, Drame, Lhoest, and Rush}]{wolf-etal-2020-transformers}
Thomas Wolf, Lysandre Debut, Victor Sanh, Julien Chaumond, Clement Delangue,
  Anthony Moi, Pierric Cistac, Tim Rault, Remi Louf, Morgan Funtowicz, Joe
  Davison, Sam Shleifer, Patrick von Platen, Clara Ma, Yacine Jernite, Julien
  Plu, Canwen Xu, Teven Le~Scao, Sylvain Gugger, Mariama Drame, Quentin Lhoest,
  and Alexander Rush. 2020.
\newblock \href {https://doi.org/10.18653/v1/2020.emnlp-demos.6} {Transformers:
  State-of-the-art natural language processing}.
\newblock In \emph{Proceedings of the 2020 Conference on Empirical Methods in
  Natural Language Processing: System Demonstrations}, pages 38--45, Online.
  Association for Computational Linguistics.

\bibitem[{Xu et~al.(2015)Xu, Callison-Burch, and
  Napoles}]{xu-etal-2015-problems}
Wei Xu, Chris Callison-Burch, and Courtney Napoles. 2015.
\newblock \href {https://doi.org/10.1162/tacl_a_00139} {Problems in current
  text simplification research: New data can help}.
\newblock \emph{Transactions of the Association for Computational Linguistics},
  3:283--297.

\bibitem[{Xu et~al.(2016)Xu, Napoles, Pavlick, Chen, and
  Callison-Burch}]{xu-etal-2016-optimizing}
Wei Xu, Courtney Napoles, Ellie Pavlick, Quanze Chen, and Chris Callison-Burch.
  2016.
\newblock \href {https://doi.org/10.1162/tacl_a_00107} {Optimizing statistical
  machine translation for text simplification}.
\newblock \emph{Transactions of the Association for Computational Linguistics},
  4:401--415.

\bibitem[{Zhang et~al.(2022{\natexlab{a}})Zhang, Li, Chen, Deng, Bi, Tan,
  Huang, and Chen}]{zhang_differentiable_2022}
Ningyu Zhang, Luoqiu Li, Xiang Chen, Shumin Deng, Zhen Bi, Chuanqi Tan, Fei
  Huang, and Huajun Chen. 2022{\natexlab{a}}.
\newblock \href {http://arxiv.org/abs/2108.13161} {Differentiable {Prompt}
  {Makes} {Pre}-trained {Language} {Models} {Better} {Few}-shot {Learners}}.
\newblock ArXiv:2108.13161 [cs].

\bibitem[{Zhang et~al.(2022{\natexlab{b}})Zhang, Yu, Shetty, Song, and
  Zhang}]{zhang-etal-2022-prompt}
Rongzhi Zhang, Yue Yu, Pranav Shetty, Le~Song, and Chao Zhang.
  2022{\natexlab{b}}.
\newblock \href {https://doi.org/10.18653/v1/2022.acl-long.55} {Prompt-based
  rule discovery and boosting for interactive weakly-supervised learning}.
\newblock In \emph{Proceedings of the 60th Annual Meeting of the Association
  for Computational Linguistics (Volume 1: Long Papers)}, pages 745--758,
  Dublin, Ireland. Association for Computational Linguistics.

\bibitem[{Zhang et~al.(2022{\natexlab{c}})Zhang, Roller, Goyal, Artetxe, Chen,
  Chen, Dewan, Diab, Li, Lin et~al.}]{zhang2022opt}
Susan Zhang, Stephen Roller, Naman Goyal, Mikel Artetxe, Moya Chen, Shuohui
  Chen, Christopher Dewan, Mona Diab, Xian Li, Xi~Victoria Lin, et~al.
  2022{\natexlab{c}}.
\newblock Opt: Open pre-trained transformer language models.
\newblock \emph{arXiv preprint arXiv:2205.01068}.

\bibitem[{Zhang et~al.(2020)Zhang, Kishore, Wu, Weinberger, and
  Artzi}]{Zhang_2020}
Tianyi Zhang, Varsha Kishore, Felix Wu, Kilian~Q. Weinberger, and Yoav Artzi.
  2020.
\newblock \href {https://openreview.net/forum?id=SkeHuCVFDr} {Bertscore:
  Evaluating text generation with {BERT}}.
\newblock In \emph{8th International Conference on Learning Representations,
  {ICLR} 2020, Addis Ababa, Ethiopia, April 26-30, 2020}. OpenReview.net.

\bibitem[{Zhang and Lapata(2017)}]{zhang-lapata-2017-sentence}
Xingxing Zhang and Mirella Lapata. 2017.
\newblock \href {https://doi.org/10.18653/v1/D17-1062} {Sentence simplification
  with deep reinforcement learning}.
\newblock In \emph{Proceedings of the 2017 Conference on Empirical Methods in
  Natural Language Processing}, pages 584--594, Copenhagen, Denmark.
  Association for Computational Linguistics.

\bibitem[{Zhu et~al.(2010)Zhu, Bernhard, and
  Gurevych}]{zhu-etal-2010-monolingual}
Zhemin Zhu, Delphine Bernhard, and Iryna Gurevych. 2010.
\newblock \href {https://aclanthology.org/C10-1152} {A monolingual tree-based
  translation model for sentence simplification}.
\newblock In \emph{Proceedings of the 23rd International Conference on
  Computational Linguistics (Coling 2010)}, pages 1353--1361, Beijing, China.
  Coling 2010 Organizing Committee.

\bibitem[{{Zhuo} et~al.(2023){Zhuo}, {Huang}, {Chen}, and {Xing}}]{zhuo2023red}
Terry~Yue {Zhuo}, Yujin {Huang}, Chunyang {Chen}, and Zhenchang {Xing}. 2023.
\newblock \href {https://doi.org/10.48550/arXiv.2301.12867} {{Red teaming
  ChatGPT via Jailbreaking: Bias, Robustness, Reliability and Toxicity}}.
\newblock \emph{arXiv e-prints}, page arXiv:2301.12867.

\end{thebibliography}
\bibliographystyle{acl_natbib}

\appendix
\newpage
% \onecolumn

\section{Model Details} \label{sec:appendix_model_details}

% \begin{table*}[t]
% \centering
%  % \setlength\tabcolsep{3pt}
%  % \def\arraystretch{1.2}
% \scalebox{0.80}{
% \begin{tabular}{lllp{7cm}}
%   \rowcolor{gray!24}
%      \textbf{\textsc{Models}} & \textbf{\textsc{Type}}& \textbf{\textsc{Variants}}  & \textbf{\textsc{Training Data}}  \\

% % \textit{Open Weight} \\
% \textsc{bloom, bloomz} & \multirow{4}{*}{D} & 560M, 1b1, 3b, 7b & ROOTS \cite{laurenccon2022bigscience}, Huggingface Datasets \cite{lhoest2021datasets} \\ 
% \textsc{LLaMa} &  & 7b, 13b, 30b, 65b  & CommonCrawl, C4 \cite{raffel2020exploring}, Github, Wikipedia, ArXiv, StackExchange \\ 
% \textsc{OPT} &  & 1.3b, 6.7b, 13b, 30b, 66b & Pile \cite{gao2020pile}, Reddit \cite{baumgartner2020pushshift}\\
% \textsc{GPT} & & 6b, 20b & Pile\\

% \addlinespace[0.3cm]

% \textsc{t5}  & \multirow{6}{*}{E-D} & base, large, small, xl, xxl & \multirow{2}{*}{C4}\\ 
% \textsc{ul2}  & & 20b & \\ 
% \textsc{t0, t0pp}  & & 3b, 11b & T0-SF \cite{sanh2022multitask} \\ 
% \textsc{flan-t5} &  & base, large, small, xl, xxl &  Muffin \cite{wei2022finetuned}, T0-SF \cite{sanh2022multitask}, NIV2 \cite{wang2022benchmarking}\\ 
% \textsc{flan-ul2}  & & 20b & Muffin, T0-SF, NIV2\\ 

% % \addlinespace[0.3cm]

% % \textit{Closed Weight} \\
% % \textsc{openai} & \multirow{1}{*}{D}  & \multirow{1}{*}{-}  & \multirow{1}{*}{-} \\
% % % \textsc{cohere} &  \\

% \end{tabular}
% }
% \caption{Description of Open-Weight models. Model type "D" refers to decoder-only models, "E-D" for models based on an encoder-decoder architecture.} \label{tab:model_stats}

% \end{table*}

\begin{table*}[t]
\centering
\scalebox{0.80}{
\begin{tabularx}{1.2\textwidth}{lcXX}
  % \rowcolor{gray!24}
\toprule
\textbf{Models} & \textbf{Type} & \textbf{Sizes}  & \textbf{Training Data}  \\
\midrule
% \textit{Open Weight} \\
\texttt{BLOOM} & D & 560M, 1b1, 3b, 7b, 175b & ROOTS \cite{laurenccon2022bigscience}, Huggingface Datasets \cite{lhoest2021datasets} \\ 
\texttt{BLOOMZ} & D & 560M, 1b1, 3b, 7b, 175b & P3 \cite{sanh2022multitask}, xP3 \\ 
\texttt{LLaMA} & D & 7b, 13b, 30b, 65b  & CommonCrawl, C4 \cite{raffel2020exploring}, Github, Wikipedia, ArXiv, StackExchange \\ 
\texttt{OPT} & D & 1.3b, 6.7b, 13b, 30b, 66b & Pile \cite{gao2020pile}, Reddit \cite{baumgartner2020pushshift}\\
\texttt{OPT-IML} & D & 1.3b, 30b, & OPT-IML Benchmark \citep{iyer2023optiml} \\
\texttt{GPT-J} & D & 6b, 20b & Pile \citep{gao2020pile} \\

% \addlinespace[0.3cm]

\texttt{T5}  & E-D & 60m (small), 220m (base), 770m (large), 3b (xl), 11b (xxl) & C4 \cite{raffel2020exploring} \\ 
\texttt{T0, T0pp}  & E-D & 3b, 11b & P3 \cite{sanh2022multitask} \\ 
\texttt{Flan-T5} & E-D & 60m (small), 220m (base), 770m (large), 3b (xl), 11b (xxl) &  Muffin \cite{wei2022finetuned}, P3 \cite{sanh2022multitask}, NIV2 \cite{wang2022benchmarking} \\ 
\texttt{UL2}  &  E-D & 20b & \\ 
\texttt{Flan-UL2}  & E-D & 20b & Muffin \cite{wei2022finetuned}, P3 \cite{sanh2022multitask}, NIV2 \cite{wang2022benchmarking} \\ 
\bottomrule
% \addlinespace[0.3cm]

% \textit{Closed Weight} \\
% \textsc{openai} & \multirow{1}{*}{D}  & \multirow{1}{*}{-}  & \multirow{1}{*}{-} \\
% % \textsc{cohere} &  \\

\end{tabularx}
}
\caption{Description of Open-Weight models. Model type "D" refers to decoder-only models, "E-D" for models based on an encoder-decoder architecture.} \label{tab:model_stats}

\end{table*}
In this section, we describe each type of LLM we use in our experiments.

\subsection{Open-weight Models}

As a brief disclaimer, we note that some listed models are not truly ``open-weight" and may require special permission to obtain weights for self-hosting. Further, in our descriptions, we do not distinguish between different variations of the same model. We provide the details of the training data and model sizes in Table~\ref{tab:model_stats}. We consider both encoder-decoder and decoder-only models for our evaluation as discussed below.

\subsubsection{Encoder-Decoder Models}
\paragraph{\texttt{T5} Family}
We evaluate a range of model variants derived from the original T5 models~\citep{raffel2020exploring}. Originally, training recipes for T5 employ pre-training with a span-infilling objective and are thus not suitable for left-to-right generation tasks off the shelf. We thus use the \texttt{T5-LM-adapted} models from \cite{lester-etal-2021-power} which have undergone continued pre-training using a standard LM objective.

One later derivation includes the instruction-tuned variant \texttt{Flan-T5}~\cite{chung2022scaling}, which continues training from the aforementioned \texttt{T5-LM-adapted} checkpoints and uses a wide variety of labeled data for instruction fine-tuning. Notably, the dataset description by \citet{chung2022scaling} does not include any reference to simplification-related tasks.
Similar parallel efforts lead to the creation of the \texttt{T0} models~\cite{sanh2022multitask}.

Finally, \texttt{UL2}~\cite{tay2023ul} proposes a more diverse set of pre-training objectives beyond simple span corruption. Additional tasks include sequence distortion and extreme span corruption.

\subsubsection{Decoder-only Models}

\paragraph{\texttt{GPT-J}/\texttt{GPT-X}}
Early reproduction efforts of large-scale GPT-style models started following the surge in popularity of GPT-2~\cite{radford2019language}. For our benchmark, we include models published by EleutherAI, namely the 6 billion parameter variant of \texttt{GPT-J} \cite{gpt-j} and the 20 billion parameter version of \texttt{GPT-NeoX}~\cite{black-etal-2022-gpt}.
Both models were trained with a standard LM pre-training objective and were not fine-tuned to follow instructions.

\paragraph{\texttt{OPT}/\texttt{LLaMA}}
Reproduction efforts of large-scale decoder-only models conducted by researchers at Meta AI were released under the \texttt{OPT} label~\cite{zhang2022opt} and more recently under the \texttt{LLaMA} label~\cite{touvron2023llama}. Besides a different composition in training data and some implementation choices relating to hardware performance, they otherwise share similar architectures and training objectives with the previously mentioned GPT-like models.
\citet{iyer2023optiml} experimented with instruction tuning the \texttt{OPT} models to provide \texttt{OPT-IML} checkpoints, which we also use in BLESS.

\paragraph{\texttt{BLOOM}}
The result of an open collaboration, the \texttt{BLOOM} model family~\cite{scao2022bloom} represents the largest open-weight models available at the time of writing, up to the full 176 billion parameter scale of \texttt{GPT-3} \citep{brown2020language}. The original model was only trained with a standard LM pre-training objective. 
\texttt{BLOOMZ} models~\cite{muennighoff2022crosslingual} extend these models with instruction fine-tuning.

\subsection{Closed-Weight Models}
%In contrast to the previous models, commercial providers do not give access to model weights, and they share details on the particular architecture rather sparingly.
%We therefore primarily report the history of their releases.

%\paragraph{OpenAI}
As the current primary choice for commercial solutions, we benchmark a range of models by OpenAI.
Previous publications regarding the GPT family~\citep{radford2018improving, radford2019language, brown2020language} establish that these models (\texttt{Ada/Babbage/Curie/Davinci}) are decoder-only, with varying numbers of parameters.
% For the best-performing variant, \texttt{GPT-3.5-Turbo}, additional instruction-tuning and more robust data filtering were incorporated. 
Table \ref{tab:costings} shows the API inference costs of our experiments with OpenAI's models.

\begin{table}[h!]
    \centering
    \small % smaller font size
    \scalebox{0.8}{
    \begin{tabularx}{1.2\columnwidth}{lXrrr}
        \toprule
        \multirow{2}{*}{Model} & \$/1k\newline tokens & \multirow{2}{*}{\asset} & \multirow{2}{*}{\medeasi} & \multirow{2}{*}{\newsela} \\
        \midrule
        \texttt{Ada-001} & 0.0004 & 0.35 & 0.41 & 0.28 \\
        \texttt{Babbage-001} & 0.0005 & 0.44 & 0.51 & 0.35 \\
        \texttt{Curie-001} & 0.002 & 1.76 & 2.01 & 1.41 \\
        \texttt{GPT-3.5-Turbo} & 0.002 & 1.75 & 1.95 & 1.37 \\
        \texttt{Davinci-002} & 0.02 & 17.62 & 20.06 & 14.10 \\
        \texttt{Davinci-003} & 0.02 & 17.52 & 19.90 & 13.96 \\
        \midrule 
        Total & -- & 39.54 & 44.84 & 31.47 \\
        \bottomrule
    \end{tabularx}
    }
    \caption{
    Pricing information for OpenAI's API models. 
    Here we report the total costs incurred for all three inference prompts and three seeded runs, totalling nine inference runs per dataset. Prices listed correspond to those for the API-based models available from April through June, 2023. 
    All prices are in USD.
    }
    \label{tab:costings}
\end{table}

% \paragraph{Cohere}
% To evaluate competing commercial models, we also run the \texttt{command-light} model by Cohere. It represents a smaller version of their RLHF-tuned instruction model and is likely comparable to the smaller OpenAI variants.

% \plan{Work in Progress}

% \section{Inference settings and infrastructure}
% \label{sec:app:hardware}
% \todo{tannon}

% Tables placing

\begin{table}[h]
\centering
\caption{Simplification Results on \asset}
\label{table:asset-results} 
\resizebox{\columnwidth}{!}{%
\begin{tabular}{@{}l|cccc}
\toprule
% \textbf{Model} &
  & \textbf{SARI$\uparrow$} &
  \textbf{FKGL$\downarrow$} & 
    \textbf{BERT$\uparrow$} &  
  \textbf{LENS$\uparrow$} \\
\midrule \midrule \midrule
\textbf{Baselines} \\
\midrule \midrule
Gold References  & 45.27  & 6.53 & 78.89 & 65.58  \\
\texttt{MUSS-mined}  & 42.29  & 8.18   & 79.86 & 61.36       \\
\texttt{MUSS-wiki-mined}  & 44.90  & 5.29  & 77.71 & 69.23       \\
\midrule \midrule \midrule
\textbf{LLMs} \\
\midrule \midrule
% cohere-command-light & 39.05 & 7.85 & 68.02 & 52 \\
\texttt{Ada-001} & 33.97 & 9.06 & 81.76 & 56.41 \\
\texttt{Babbage-001} & 38.44 & 8.65 & 82.46 & 61.39 \\
\texttt{Curie-001} & 39.87 & 8.33 & 82.75 & 63.02 \\
\texttt{Davinci-002} & 42.84 & 7.77 & \textbf{85.91} & 67.09 \\
\texttt{Davinci-003} & 46.6 & 7.74 & 79.66 & 67.39 \\
\midrule
\texttt{GPT-3.5-Turbo} & \textbf{47.69} & 7.51  & 79.39 & \textbf{69.17} \\
\midrule
\texttt{BLOOM-560m} & 36.14 & 8.01 & 50.11 & 42.68 \\
\texttt{BLOOM-1b1} & 34.08 & 8.18 & 68.60 & 51.23 \\
\texttt{BLOOM-3b} & 37.15 & 7.92 & 72.28 & 54.34 \\
\texttt{BLOOM-7b1} & 36.96 & 8.17 & 77.82 & 57.37 \\
\texttt{BLOOM} & 39.72 & 7.78 & 76.63 & 60.37 \\
\midrule
\texttt{BLOOMZ-560m} & 35.12 & 7.52 & 41.21 & 39.52 \\
\texttt{BLOOMZ-1b1} & 35.00 & 8.42 & 76.66 & 54.86 \\
\texttt{BLOOMZ-3b} & 35.74 & 8.73 & 75.86 & 56.78 \\
\texttt{BLOOMZ-7b1} & 37.05 & 8.56 & 79.09 & 59.14 \\
\texttt{BLOOMZ} & 37.63 & 8.27 & 82.06 & 61.07 \\
\midrule
\texttt{GPT-J-6b} & 38.86 & 7.83 & 76.48 & 60.13 \\
\texttt{GPT-NeoX-20b} & 39.04 & 8.04 & 75.81 & 60.87 \\
\midrule
\texttt{LLaMA-7b} & 40.70 & 7.39 & 75.52 & 62.80 \\
\texttt{LLaMA-13b} & 40.45 & 7.33 & 76.13 & 62.95 \\
\texttt{LLaMA-30b} & 39.14 & 7.32 & 78.74 & 62.73 \\
\texttt{LLaMA-65b} & 38.59 & 8.07 & 81.59 & 62.90 \\
\midrule
\texttt{OPT-1.3b} & 33.01 & 8.61 & 75.57 & 57.08 \\
\texttt{OPT-6.7b} & 38.64 & 7.79 & 76.62 & 61.26 \\
\texttt{OPT-13b} & 38.78 & 8.03 & 79.08 & 60.51 \\
\texttt{OPT-30b} & 38.04 & 8.17 & 77.22 & 60.01 \\
\texttt{OPT-66b} & 39.64 & 7.76 & 76.72 & 61.68 \\
\midrule
\texttt{OPT-IML-Max-1.3b} & 36.00 & 7.66 & 79.73 & 61.31 \\
\texttt{OPT-IML-Max-30b} & 42.03 & 6.62 & 79.39 & 65.29 \\
\midrule
\texttt{T0-3b} & 35.16 & 8.90 & 54.92 & 50.38 \\
\texttt{T0} & 36.49 & 8.56 & 55.32 & 48.71 \\
\texttt{T0pp} & 35.05 & 8.65 & 47.69 & 44.67 \\
\midrule
\texttt{T5-small-LM-adapt} & 33.89 & \textbf{6.61} & 10.27 & 14.56 \\
\texttt{T5-base-LM-adapt} & 34.70 & 6.80 & 19.63 & 14.27 \\
\texttt{T5-large-LM-adapt} & 31.12 & 6.88 & 37.82 & 15.21 \\
\texttt{T5-xl-LM-adapt} & 29.12 & 7.06 & 48.25 & 23.39 \\
\texttt{T5-xxl-LM-adapt} & 33.17 & 6.85 & 46.59 & 25.43 \\
\midrule
\texttt{Flan-T5-small} & 38.57 & 7.58 & 77.26 & 54.80 \\
\texttt{Flan-T5-base} & 41.40 & 7.32 & 79.70 & 62.75 \\
\texttt{Flan-T5-large} & 42.17 & 6.78 & 80.44 & 63.35 \\
\texttt{Flan-T5-xl} & 41.07 & 7.16 & 85.06 & 64.74 \\
\texttt{Flan-T5-xxl} & 41.75 & 7.27 & 84.13 & 66.08 \\
\midrule
\texttt{UL2} & 35.65 & 7.65 & 37.01 & 15.99 \\
\texttt{Flan-UL2} & 42.83 & 6.85 & 84.34 & 67.36 \\

\bottomrule
\end{tabular}
}
\end{table}

\begin{table}[h]
\centering
\caption{Simplification Results on \medeasi}
\label{table:med-easi-results} 
\resizebox{\columnwidth}{!}{%
\begin{tabular}{@{}l|cccc}
\toprule
\textbf{Model} &
  \textbf{SARI$\uparrow$} &
  \textbf{FKGL$\downarrow$} & 
    \textbf{BERT$\uparrow$} &  
  \textbf{LENS$\uparrow$} \\
\midrule \midrule \midrule
\textbf{Baselines} \\
\midrule \midrule
Gold References  & 100 & 9.59 & 100 & 65.89 \\
\texttt{MUSS-mined} & 35.15 & 9.29 & 42.55 & 52.48 \\
\texttt{MUSS-wiki-mined} & 35.12 & 8.04 & 43.07 & 59.12 \\
\midrule \midrule \midrule
\textbf{LLMs} \\
\midrule \midrule
% cohere-command-light & -- & -- & -- & --\\
\texttt{Ada-001} & 36.52 & 10.62 & 33.95 & 41.43 \\
\texttt{Babbage-001} & 36.60 & 10.49 & 37.95 & 53.91 \\
\texttt{Curie-001} & 38.22 & 10.15 & 39.31 & 56.10 \\
\texttt{Davinci-002} & 36.34 & 10.05 & 43.67 & 57.71 \\
\texttt{Davinci-003} & 39.81 & \textbf{9.31} & 40.83 & 60.71 \\
\midrule
\texttt{GPT-3.5-Turbo} & \textbf{40.14} & 8.93 & 40.67 & \textbf{63.80} \\
\midrule
\texttt{BLOOM-560m} & 35.37 & 7.58 & -2.60 & 36.27 \\
\texttt{BLOOM-1b1} & 35.86 & 7.37 & 1.63 & 40.47 \\
\texttt{BLOOM-3b} & 35.48 & 7.40 & 5.94 & 42.21 \\
\texttt{BLOOM-7b1} & 37.47 & 7.23 & 9.53 & 44.17 \\
\texttt{BLOOM} & 37.72 & 7.11 & 11.95 & 47.50 \\
\midrule
\texttt{BLOOMZ-560m} & 33.14 & 6.83 & -3.08 & 38.32 \\
\texttt{BLOOMZ-1b1} & 35.65 & 6.99 & 6.40 & 43.69 \\
\texttt{BLOOMZ-3b} & 35.68 & 7.17 & 8.56 & 44.79 \\
\texttt{BLOOMZ-7b1} & 36.78 & 7.08 & 9.43 & 47.15  \\
\texttt{BLOOMZ} & 36.60 & 7.08 & 12.90 & 47.67 \\
\midrule
\texttt{GPT-J-6b} & 36.20 & 7.01 & 10.53 & 46.67 \\
\texttt{GPT-NeoX-20b} & 36.02 & 7.07 & 10.62 & 46.46 \\
\midrule
\texttt{LLaMA-7b} & 36.95 & 6.62 & 10.28 & 48.42 \\
\texttt{LLaMA-13b} & 36.98 & 6.73 & 11.43 & 48.63  \\
\texttt{LLaMA-30b} & 37.56 & 6.89 & 12.21 & 47.92 \\
\texttt{LLaMA-65b} & 37.86 & 6.85 & 12.20 & 47.45 \\
\midrule
\texttt{OPT-1.3b} & 34.00 & 7.17 & 3.82 & 43.64 \\
\texttt{OPT-6.7b} & 34.73 & 7.02 & 8.86 & 47.72 \\
\texttt{OPT-13b} & 34.69 & 6.96 & 8.73 & 47.16 \\
\texttt{OPT-30b} & 35.08 & 7.02 & 9.96 & 46.96 \\
\texttt{OPT-66b} & 35.72 & 6.96 & 11.42 & 47.28 \\
\midrule
\texttt{OPT-IML-Max-1.3b} & 37.01 & 7.12 & 11.85 & 46.80 \\
\texttt{OPT-IML-Max-30b} & 35.80 & 6.78 & 11.73 & 49.28 \\
\midrule
\texttt{T0-3b} & 38.16 & 10.34 & 17.83 & 42.02  \\
\texttt{T0} & 35.67 & 10.81 & 15.93 & 42.76 \\
\texttt{T0pp} & 35.61 & 10.67 & 11.58 & 36.60 \\
\midrule
\texttt{T5-small-LM-adapt} & 34.71 & 8.87 & -4.15 & 12.68 \\
\texttt{T5-base-LM-adapt} & 34.70 & 8.47 & -0.92 & 16.41 \\
\texttt{T5-large-LM-adapt} & 36.69 & 8.62 & 10.27 & 19.34 \\
\texttt{T5-xl-LM-adapt} & 33.65 & 8.83 & 18.91 & 22.59 \\
\texttt{T5-xxl-LM-adapt} & 32.61 & 9.10 & 21.69 & 28.21 \\
\midrule
\texttt{Flan-T5-small} & 36.65 & 8.99 & 38.60 & 45.37 \\
\texttt{Flan-T5-base} & 36.79 & 9.05 & 40.63 & 51.95 \\
\texttt{Flan-T5-large} & 35.71 & 8.70 & 41.31 & 52.59 \\
\texttt{Flan-T5-xl} & 33.21 & 9.11 & \textbf{44.12} & 54.75 \\
\texttt{Flan-T5-xxl} & 34.27 & 9.13 & 43.43 & 54.70 \\
\midrule
\texttt{UL2} & 35.89 & 9.28 & 17.15 & 19.79 \\
\texttt{Flan-UL2} & 35.31 & 8.52 & 42.80 & 57.95 \\

\bottomrule
\end{tabular}
}
\end{table}

\begin{table}[h]
\centering
\caption{Simplification Results on \newsela}
\label{table:newsela-results} 
\resizebox{\columnwidth}{!}{%
\begin{tabular}{@{}l|cccc}
\toprule
\textbf{Model} &
  \textbf{SARI$\uparrow$} &
  \textbf{FKGL$\downarrow$} & 
    \textbf{BERT$\uparrow$} &  
  \textbf{LENS$\uparrow$} \\
\midrule \midrule \midrule
\textbf{Baselines} \\
\midrule \midrule
Gold References  & 60.11  & 5.88 & 87.66 & 71.02  \\
\texttt{MUSS-mined} & 38.40 & 7.86 & 72.14 & 61.49 \\
\texttt{MUSS-wiki-mined} & \textbf{41.24} & 6.12 & 74.10 & 67.61 \\
\midrule \midrule \midrule
\textbf{LLMs} \\
\midrule \midrule
% cohere-command-light \\
\texttt{Ada-001} & 34.42 & 8.66 & 70.33 & 55.06 \\
\texttt{Babbage-001} & 36.41 & 8.32 & 62.99 & 60.91 \\
\texttt{Curie-001} & 37.53 & 8.23 & 69.17 & 64.35 \\
\texttt{Davinci-002} & 40.25 & 7.46 & 73.62 & \textbf{68.58} \\
\texttt{Davinci-003} & 37.76 & 7.75 & 61.56 & 66.20 \\
\midrule
\texttt{GPT-3.5-Turbo} & 37.29 & 7.80 & 60.19 & 67.97 \\
\midrule
\texttt{BLOOM-560m} & 33.41 & 7.76 & 31.85 & 38.58 \\
\texttt{BLOOM-1b1} & 35.37 & 7.99 & 48.52 & 46.54   \\
\texttt{BLOOM-3b} & 35.85 & 8.22 & 55.33 & 51.71 \\
\texttt{BLOOM-7b1} & 36.12 & 7.96 & 61.00 & 54.16 \\
\texttt{BLOOM} & 37.48 & 7.49 & 61.17 & 60.98 \\
\midrule
\texttt{BLOOMZ-560m} & 28.55 & 7.53 & 17.56 & 34.21 \\
\texttt{BLOOMZ-1b1} & 35.22 & 7.47 & 54.19 & 53.05 \\
\texttt{BLOOMZ-3b} & 34.75 & 8.51 & 52.51 & 52.37 \\
\texttt{BLOOMZ-7b1} & 36.21 & 8.29 & 59.53 & 59.36 \\
\texttt{BLOOMZ} & 37.06 & 8.41 & 69.55 & 62.07 \\
\midrule
\texttt{GPT-J-6b} & 36.8 & 7.47 & 59.59 & 58.98 \\
\texttt{GPT-NeoX-20b} & 36.87 & 7.62 & 56.85 & 59.71 \\
\midrule
\texttt{LLaMA-7b} & 36.70 & 6.28 & 55.31 & 62.43 \\
\texttt{LLaMA-13b} &  37.16 & 6.42 & 59.61 & 63.32 \\
\texttt{LLaMA-30b} & 37.50 & 6.75 & 63.89 & 64.30 \\
\texttt{LLaMA-65b} & 38.59 & 7.10 & 67.82 & 64.24 \\
\midrule
\texttt{OPT-1.3b} & 34.76 & 7.96 & 50.78 & 55.35 \\
\texttt{OPT-6.7b} & 36.58 & 7.76 & 58.68 & 60.28 \\
\texttt{OPT-13b} & 37.67 & 7.16 & 60.65 & 61.31 \\
\texttt{OPT-30b} & 37.58 & 7.75 & 61.79 & 61.91 \\
\texttt{OPT-66b} & 37.45 & 7.25 & 60.43 & 62.98 \\
\midrule
\texttt{OPT-IML-Max-1.3b} & 37.08 & 7.32 & 62.68 & 60.47 \\
\texttt{OPT-IML-Max-30b} & 39.59 & 6.09 & 66.39 & 64.74 \\
\midrule
\texttt{T0-3b} & 33.37 & 8.50 & 36.56 & 50.64 \\
\texttt{T0} & 32.83 & 7.58 & 30.96 & 53.23 \\
\texttt{T0pp} & 33.02 & 8.20 & 30.66 & 47.62 \\
\midrule
\texttt{T5-small-LM-adapt} & 30.54 & 6.33 & 4.85 & 16.85 \\
\texttt{T5-base-LM-adapt} & 32.94 & \textbf{6.00} & 12.13 & 17.80 \\
\texttt{T5-large-LM-adapt} & 33.48 & 6.35 & 30.99 & 20.07 \\
\texttt{T5-xl-LM-adapt} & 32.85 & 6.62 & 42.51 & 25.48 \\
\texttt{T5-xxl-LM-adapt} & 33.44 & 6.67 & 44.15 & 29.49 \\
\midrule
\texttt{Flan-T5-small} & 37.72 & 7.61 & 68.15 & 53.61 \\
\texttt{Flan-T5-base} & 38.67 & 7.21 & 68.09 & 59.11 \\
\texttt{Flan-T5-large} & 39.08 & 6.90 & 70.27 & 62.70 \\
\texttt{Flan-T5-xl} & 37.51 & 7.25 & \textbf{75.50} & 64.39 \\
\texttt{Flan-T5-xxl} & 39.42 & 7.32 & 73.05 & 65.13 \\
\midrule
\texttt{UL2} & 35.22 & 6.92 & 37.07 & 21.62  \\
\texttt{Flan-UL2} & 40.27 & 6.86 & 73.23 & 66.42 \\

\bottomrule
\end{tabular}
}
\end{table}

\section{Supplemental Results} 
\label{sec:appendix_more_results}

Tables \ref{table:asset-results}, \ref{table:med-easi-results}, and \ref{table:newsela-results} show full results for these on \asset, \medeasi, and \newsela respectively.

\subsection{Details on Evaluation Metrics}
\label{sec:sup_eval_metrics}

% \label{sec:appendix_eval}
A variety of automatic evaluation methods have been proposed. Commonly used automatic metrics like BLEU \citep{papineni-etal-2002-bleu} and SARI \citep{xu-etal-2016-optimizing} can provide insights into how similar a model's outputs are to a set of gold reference simplifications. However, to more precisely understand a model's strengths and weaknesses, finer-grained evaluation is often required. For example, calculating the distribution of edit simplification operations (e.g.\ additions and deletions) \citep{vasquez-rodriguez-etal-2021-investigating, Vasquez-Rodriguez-2021b} can yield more insights into the capabilities of these systems. 
% Human evaluation has also been performed \citep{xu-etal-2016-optimizing}. 
% For example, previous work has investigated the trade-off between simplicity and meaning preservation \citep{Schwarzer2018}. 
We evaluate model outputs according to multiple metrics. 
While we focus on reporting SARI and BERTScore in order to relate our findings with previous work, we also compute additional evaluation metrics for more fine-grained analyses and perform a qualitative analysis.
Specifically, we report:

\begin{enumerate}
        \item \textbf{SARI} \citep{xu-etal-2016-optimizing}: SARI (\textbf{S}ystem output \textbf{A}gainst \textbf{R}eferences and against the \textbf{I}nput sentence) is a holistic metric for simplification quality. It computes the F1 score for n-grams added, kept, and deleted, with respect to the input (source) and reference sentences.
        \item \textbf{BERTScore} \citep{Zhang_2020}: We compute the BERTScore precision, recall and F1 of the predictions against both the reference and source sentences, totaling in 6 different scores. Results reported in the paper use BERTScore F1 computed between system output simplifications and the gold reference sentence(s).
        \item \textbf{FKGL} \citep{Kincaid-1975}: FKGL (\textbf{F}lesch-\textbf{K}incaid \textbf{G}rade \textbf{L}evel) is a weighted score based on sentence length and syllable information. The lower the FKGL, the simpler the output, and the lowest possible score is -3.40. However, for a given test set, we consider the best FKGL to be the score that is closest to the FKGL of the gold references.
        \item \textbf{LENS} \cite{maddela-etal-2023-lens}: LENS (\textbf{L}earnable \textbf{E}valuation \textbf{M}etric for \textbf{T}ext \textbf{S}implification) is a score between 0 and 100 estimated by a model trained on complex-simple pairs annotated with human ratings. We report the average LENS score for each dataset.
\end{enumerate}

% \paragraph{Prompt strategies}
% In order to gauge the impact of different prompt strategies on performance, we consider the pairwise relationships among SARI, BERTScore, FKGL, and LENS for all prompts and models on ASSET.
% This is depicted in Figure \ref{fig:all_prompts_scatter} and emphasizes that prompt 0 and prompt 2 are on par, however, prompt 0 results in slightly higher SARI scores for more models, while prompt 2, which explicitly encourages more transformative simplification operations is punished by SARI.

\begin{figure*}[h]
\centering
\includegraphics[width=0.95\linewidth]{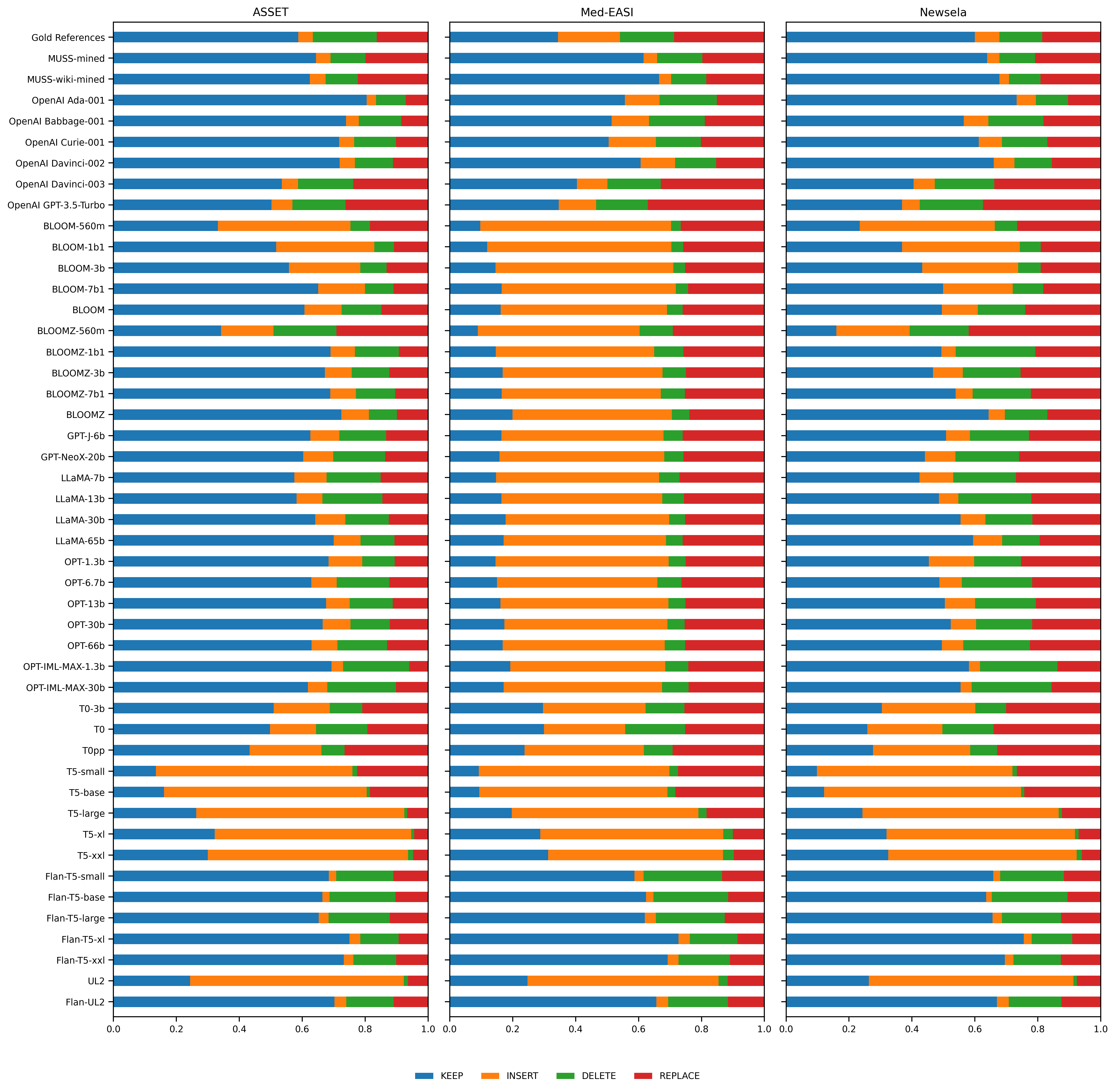}
\caption{Token-level edit operations computed for all models and test sets using prompt 2.
% Figure \ref{fig:edit_ops} shows the edit operations performed by all models on all test datasets. 
For most models, the edit operations performed in \asset and \newsela reflect those in the gold reference simplifications.
However, on the \medeasi dataset, we observe a sudden spike in insertions from all LLMs except for OpenAI and Flan models. These additions indicate the presence of potentially unrelated hallucinated tokens and endless generations, which aligns with the low BERTScore results.
We regard this failure case to be related to the fact that \medeasi presents a challenging domain which is out of the distribution of most general-purpose models.
}
\label{fig:edit_ops}
\end{figure*}

% , with the exception of outputs from the medical domain, where 
% models from the \texttt{OPT}, \texttt{LLaMA}, and \texttt{BLOOM} families show a high level of diversity

\begin{figure*}[h]
\centering
\includegraphics[width=0.9\linewidth]{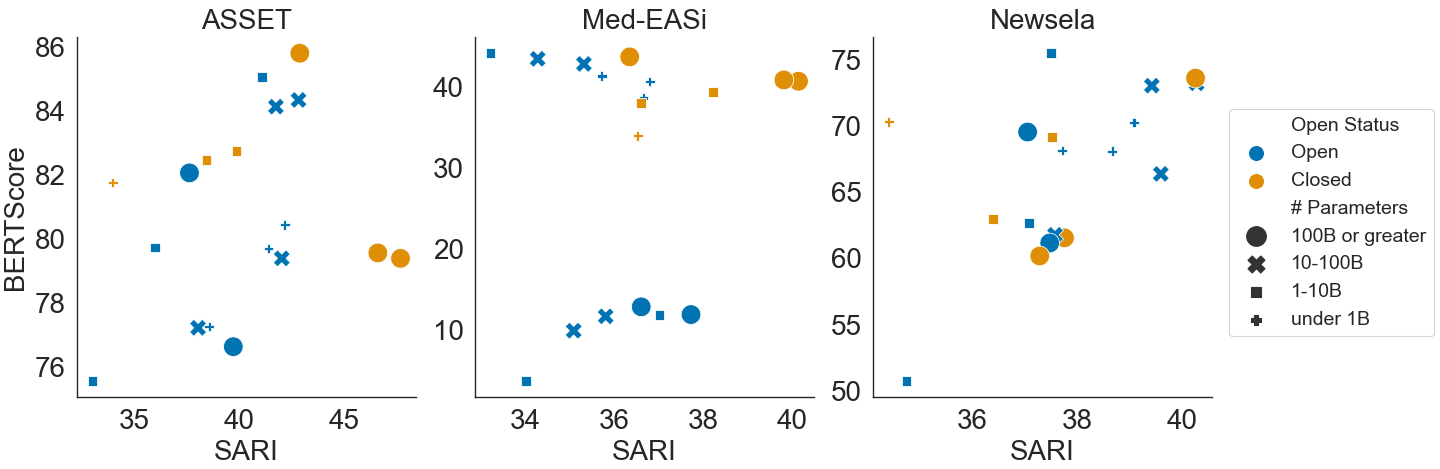}

\caption{Adequacy-simplicity trade-off as exhibited by a limited set of models on each of the three datasets. 
On \asset, higher SARI is associated with lower BERTScore. 
In the case of \medeasi, we can see that smaller models, which often tend to copy the input sentence, are rewarded by BERTScore but punished by SARI. Here, only the closed-weight OpenAI models exhibit a favorable balance between the two metrics.
On \newsela, the relationship is more linear. We suspect that this is influenced by the fact that reference sentences are taken from multiple simplification levels (1-4) and therefore cover a broader range of possible rewrites, some with more simplifying edit operations (rewarded by SARI) and some with fewer (rewarded by BERTScore).
}
\label{fig:sari_vs_bert}
\end{figure*}

% \begin{figure*}[h]
% \centering
% \begin{subfigure}[b]{.45\linewidth}
% \includegraphics[width=\linewidth]{figs/sari_vs_bert/asset_sari_F1_bert_ref_scatter_all.png}
% \caption{\asset}\label{fig:sari_vs_bert_sim_asset}
% \end{subfigure}

% \begin{subfigure}[b]{.45\linewidth}
% \includegraphics[width=\linewidth]{figs/sari_vs_bert/med_easi_sari_F1_bert_ref_scatter_all.png}
% \caption{\medeasi}\label{fig:sari_vs_bert_sim_medeasi}
% \end{subfigure}
% \begin{subfigure}[b]{.45\linewidth}
% \includegraphics[width=\linewidth]{figs/sari_vs_bert/newsela_sari_F1_bert_ref_scatter_all.png}
% \caption{\newsela}\label{fig:sari_vs_bert_sim_newsela}
% \end{subfigure}

% \caption{Comparison of SARI with BERTScore}
% \label{fig:sari_vs_bert}
% \end{figure*}

\begin{figure*}[h]
\centering
\includegraphics[width=0.95\linewidth]{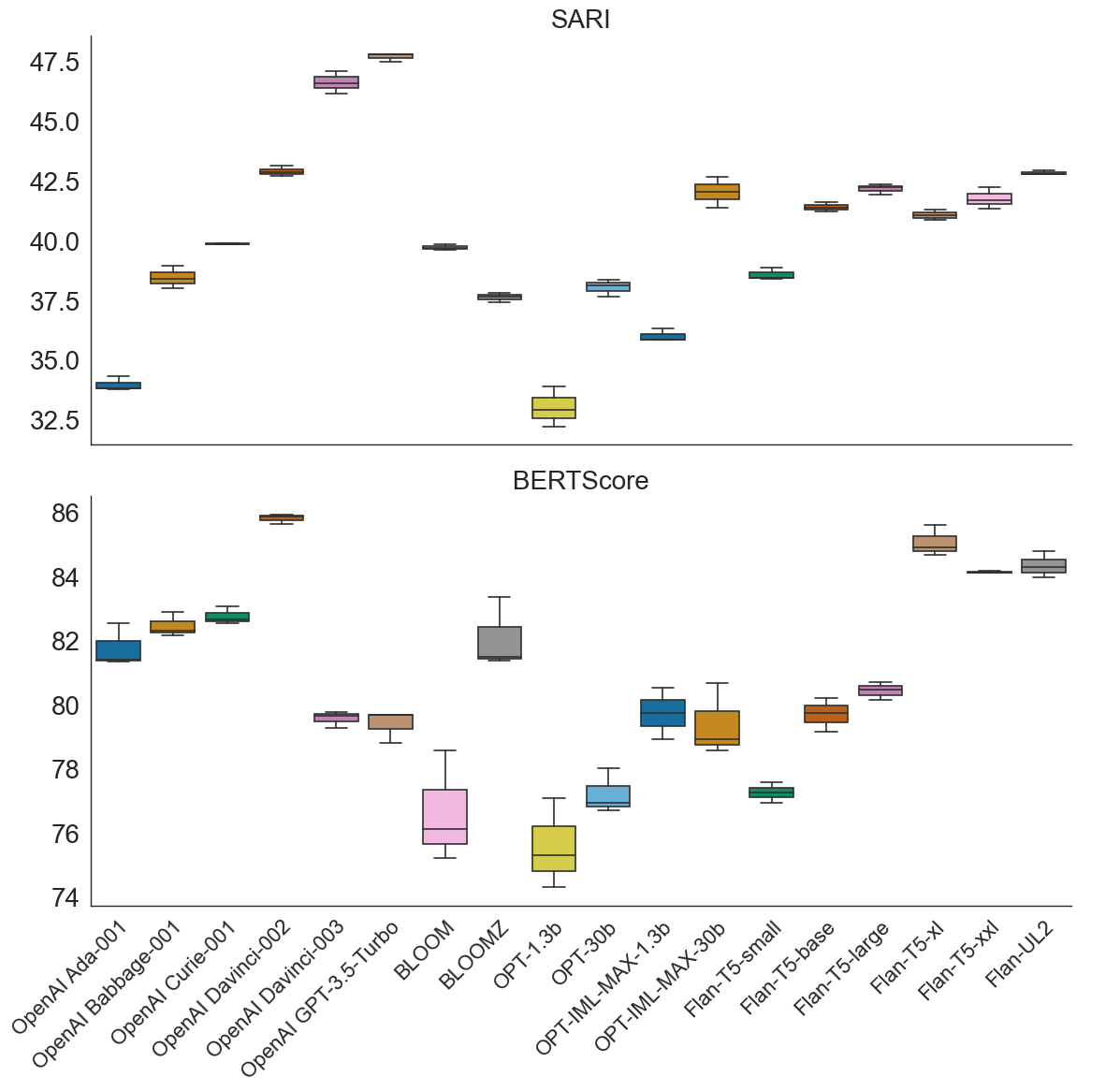}
\caption{Visualizing LLM performance for select models, generated using prompt 2. This visualization corresponds to the results reported in Table \ref{table:results-summary} for \asset. Models on the x-axis are ordered by model family, and within each model family, they are ordered by size (ascending).}
\label{fig:sari_bert_boxplots}
\end{figure*}

\begin{table*}[tbh]
\centering
\scalebox{0.90}{
\begin{tabular}{lp{10cm}c}
\toprule
         \textbf{Model} & \textbf{Sentence} & \textbf{Annotation} \\
\midrule
    Complex & They are rivaled as biological materials in toughness only by chitin. &  - \\
    Reference & They are rivaled only by chitin in toughness. & - \\
    \texttt{GPT-3.5-Turbo} & Chitin is the only biological material that \textbf{rivals} them in \textbf{toughness}. &  S↑ P+ L+ \\
    \texttt{Davinci-003} & Chitin is the only biological material \textbf{tougher} than them. & S↑ P+ L+\\
    \texttt{Davinci-002} & They are tough like chitin, which is the \textbf{toughest} \textcolor{red}{known} biological material. & MP$\downarrow$ P+ L+ \\
    \texttt{Flan-UL2} & They are only \textbf{second} to chitin for biological materials. & MP$\downarrow$ P+ L+\\
    \texttt{Flan-T5-large} & Chitin is better than \textcolor{red}{human} materials in \textbf{toughness}. & MP$\downarrow$ P+ L+ \\
\bottomrule
\end{tabular}
}
\caption{Annotation examples from a SARI-based model ranking. S: Simplification, P: Paraphrasing, L: Lexical Simplification, and MP: meaning preservation. We highlight lexical simplification in \textbf{bold} and conflicts in meaning preservation in \textcolor{red}{red}.} 
\label{tab:annotation_scheme_sari_example}
\end{table*}

\end{document}